%% file: Template_combine.tex
\documentclass{article}
\usepackage{spconf,amsmath,graphicx,hyperref,amsthm,amsfonts}
\usepackage{multirow}
\usepackage{xcolor}

\usepackage{adjustbox}
\usepackage{tabularray}
\usepackage{amssymb}

\input{preamble}

\input{symbol}

\title{Sonicmesh: Enhancing 3D Human Mesh Reconstruction in
Vision-Impaired Environments With Acoustic Signals}
\name{Xiaoxuan Liang$^1$, Hong Zhou$^2$, Zhaolong Wei$^1$, Yansong Li$^{3,4}$, Shujian Yu$^5$, Jeremy Gummeson$^1$\thanks{Corresponding author: jgummeso@umass.edu (J.~Gummeson).}}

\address{$^1$UMass, Amherst \; $^2$Devz AI \; $^3$Eastern Institute of Technology, Ningbo \; $^4$UIC \; $^5$VU Amsterdam}
\begin{document}
%\ninept
%
\maketitle
\begin{abstract}
3D human mesh reconstruction (HMR) from RGB images often degrades under poor illumination, occlusion, and non-line-of-sight conditions. Acoustic sensing provides complementary spatial cues but suffers from
low spatial resolution. We propose \textbf{SonicMesh}, which, to the best of our knowledge, is the first acoustic--visual framework for robust 3D human mesh reconstruction. SonicMesh first converts ultrasonic echoes into range--azimuth acoustic images through an Inverse Synthetic Aperture Radar (ISAR)-based imaging process. It then introduces a cross-dimensional anatomical registration module that maps modality-specific 2D joint features into a common canonical 3D human space. The registered anatomical representations are further integrated with acoustic and visual features through a two-stage fusion network for final mesh reconstruction. Experiments demonstrate that SonicMesh achieves accurate and robust 3D human reconstruction across normal, poor-light, occluded, and non-line-of-sight environments, consistently outperforming existing RGB-, radio-frequency (RF)-, and mmWave-based approaches under challenging sensing conditions.
\end{abstract}

\section{Introduction}
\label{sec:intro}

%Human mesh reconstruction (HMR) from fragmented RGB data is challenging. Data can be fragmented in scenarios such as occlusions, where parts of the body are hidden from view, and variations in body poses and shapes across individuals. Additional complications include the complexity of clothing and accessories, as well as the inherent ambiguity in inferring depth information from 2D images in computer vision. Without a complete 3D mesh, applications like AR/VR~\cite{tang2021towards}, human pose estimation~\cite{zheng2022multi}, and human-computer interaction~\cite{ryselis2020multiple} become infeasible. Recent works have explored alternative sensing technologies to enhance camera capabilities~\cite{zhao2019through}, including mmWave~\cite{xue2021mmmesh,chen2022mmbody}, Radio Frequency (RF)~\cite{geng2022densepose,zhao2019through}, and LiDAR~\cite{li2022deepfusion,bai2022transfusion}. However, these sensing methods have limitations, such as high costs, specialized hardware requirements, and significant power consumption.

Human mesh reconstruction (HMR) from RGB images remains challenging under adverse visual conditions such as poor illumination, occlusion, and non-line-of-sight (NLOS) settings, where body observations can become incomplete or unreliable. In addition, monocular RGB images inherently suffer from depth ambiguity, further complicating accurate 3D body reconstruction. Reliable HMR is important for applications such as AR/VR~\cite{tang2021towards}, human pose estimation~\cite{zheng2022multi}, and human--computer interaction~\cite{ryselis2020multiple}. To improve robustness beyond purely visual sensing, recent studies have explored alternative modalities, including mmWave~\cite{xue2021mmmesh,chen2022mmbody}, radio-frequency (RF) sensing~\cite{geng2022densepose,zhao2019through}, and LiDAR~\cite{li2022deepfusion,bai2022transfusion}.
Although these sensing technologies provide complementary geometric or depth information, they often rely on specialized hardware and can incur additional deployment cost or power consumption.

% In contrast, acoustic signals provide millimeter-level positioning accuracy~\cite{li2020fm}, cost-effectiveness, more privacy guarantee, ease of deployment, and capability of penetrating obstacles. 
% Techniques such as Synthetic Aperture Radar (SAR)~\cite{mao2018aim} use acoustic signals for object imaging, while others utilize phase changes in echoes for gesture recognition~\cite{li2020fm}. 
% Therefore, Acoustic signals offer a promising alternative for HMR using only RGB images. 

In contrast, acoustic sensing offers an attractive alternative due to its low cost, ease of deployment, privacy preservation, and robustness to adverse visual conditions~\cite{liang2024sondar,xu2025acoustic}. Owing to the relatively slow propagation speed of sound, acoustic signals can also provide fine-grained ranging information, with millimeter-level positioning accuracy~\cite{li2020fm}. Existing studies have demonstrated the potential of acoustic sensing for spatial perception, including acoustic imaging~\cite{mao2018aim} and echo-based tracking or gesture recognition~\cite{li2020fm}. These properties make acoustic sensing a promising complementary modality for robust HMR when RGB observations are incomplete or unreliable.

%However, to the best of our knowledge, no existing methods have successfully reconstructed a complete 3D mesh of the human body using only acoustic signals due to challenges such as environmental noise, signal attenuation, and weak echo signals. Also, images generated by acoustic signals~\cite{liang2024sondar} have a low resolution.

To the best of our knowledge, existing acoustic sensing methods have not yet demonstrated full 3D human mesh reconstruction, primarily due to environmental noise, signal attenuation, weak reflections, and the limited spatial resolution of acoustic images~\cite{liang2024sondar}. Motivated by these limitations, we propose \textbf{SonicMesh}, which, to the best of our knowledge, is the first end-to-end acoustic--visual framework for full 3D human mesh reconstruction. SonicMesh converts ultrasonic echoes into range--azimuth acoustic images, aligns acoustic and visual joint features in a common 3D anatomical space, and then performs global--local multimodal fusion for mesh reconstruction. In summary, our contributions include:
\begin{itemize}
\item We propose \textbf{SonicMesh}, the first acoustic--visual framework for full-body 3D human mesh reconstruction using commodity acoustic sensing.

\item We transform ultrasonic echoes into range--azimuth acoustic images and design a cross-dimensional anatomical registration mechanism that aligns acoustic and visual  joint features in a shared common 3D space.

\item We develop a two-stage acoustic--visual fusion framework that integrates global body structure and local anatomy-aware features, enabling robust HMR under poor illumination, occlusion, and NLOS conditions.
\end{itemize}

% \section{Related Work}

% \subsection{Information Bottleneck (IB) in Multimodal Learning}

% ...

\section{Proposed Method}

\subsection{Acoustic Human-Body Imaging}
\label{sec:acoustic_imaging}

%Unlike RGB cameras, which capture the appearance of a human body through light reflection, 

Acoustic sensing characterizes the body through sound reflections from different anatomical regions. We therefore first transform the received acoustic echoes into a two-dimensional spatial representation, referred to as an \emph{acoustic image}. Specifically, we adopt an Inverse Synthetic Aperture Radar (ISAR)-based acoustic imaging scheme~\cite{liang2024sondar}, where the human subject moves relative to a stationary speaker–microphone pair, forming a synthetic aperture across consecutive chirps, while the body is modeled as a collection of acoustic scatterers.

We transmit a sequence of linear frequency-modulated (LFM) chirps. For a chirp duration $T_c$, carrier frequency $f_c$, and bandwidth $B$, the transmitted signal is given by:
\begin{equation}
s(t)
=
\operatorname{rect}\!\left(\frac{t}{T_c}\right)
\cos\!\left[
2\pi
\left(
f_c t + \frac{k}{2}t^2
\right)
\right],
\qquad
k=\frac{B}{T_c},
\label{eq:chirp}
\end{equation}
where $k$ denotes the chirp rate and $\operatorname{rect}(\cdot)$ is the rectangular window function. In our implementation, the transmitted frequency sweeps from
18~kHz to 22~kHz.

%which suppresses most environmental acoustic interference while retaining reflections from the human body.

The transmitted chirp is reflected by the human body and captured by a microphone. For an echo with propagation delay $\Delta\tau$, its corresponding propagation distance is $D = \Delta\tau \times v_s /2$,
where $v_s$ is the sound speed in air. Different body parts therefore produce reflections with different delays and amplitudes, providing spatial cues for human-body imaging.

The received echoes are organized into an $M\times N$ profile matrix, where each of the $N$ consecutive profiles contains $M$ samples along the range dimension. We then apply the Hilbert transform and dechirp each profile to
obtain the intermediate-frequency (IF) matrix, followed by a 2D FFT along the range and azimuth dimensions, mapping each physical scatterer to a corresponding response in the acoustic image $I^{a}$.

%We then apply the Hilbert transform to the received echoes, followed by dechirping each column to obtain the intermediate-frequency (IF) signal matrix. Subsequently, a two-dimensional Fast Fourier Transform (2D FFT) is performed along the range and azimuth dimensions, mapping each physical scatterer to a corresponding response in the acoustic image $I^{a}$.

For a scatterer located at $(p_r,p_a)$, its response in the range--azimuth acoustic image can be expressed as:
\begin{equation}
I^{a}(i_a,i_r)
=
A
\,\operatorname{sinc}
\left[
\frac{2kT}{v_s}(i_r-p_r)
\right]
\operatorname{sinc}
\left[
\frac{2\theta}{\lambda}(i_a-p_a)
\right],
\label{eq:acoustic_image}
\end{equation}
where $A$ is a constant determined by the signal amplitude and chirp duration, $v_s$ denotes the speed of sound, $\lambda$ is the acoustic wavelength, $\theta$ denotes the effective target rotation angle during imaging, and $i_r$ and $i_a$ are the indices along the range and azimuth dimensions, respectively. Additional details are provided in the Appendix~\ref{app:acoustic_details}.

%Compared with an RGB image $I^{v}$, the acoustic image $I^{a}$ contains weaker texture and boundary information due to the much longer wavelength of acoustic signals. Nevertheless, it provides complementary spatial cues and remains informative under poor illumination and visual occlusion. We therefore treat $I^{a}$ and $I^{v}$ as two complementary observations of the same human body and subsequently map their modality-specific features into a shared canonical 3D anatomical space.

\begin{figure*}[t]
            \centering
            \includegraphics[width=0.9\linewidth]{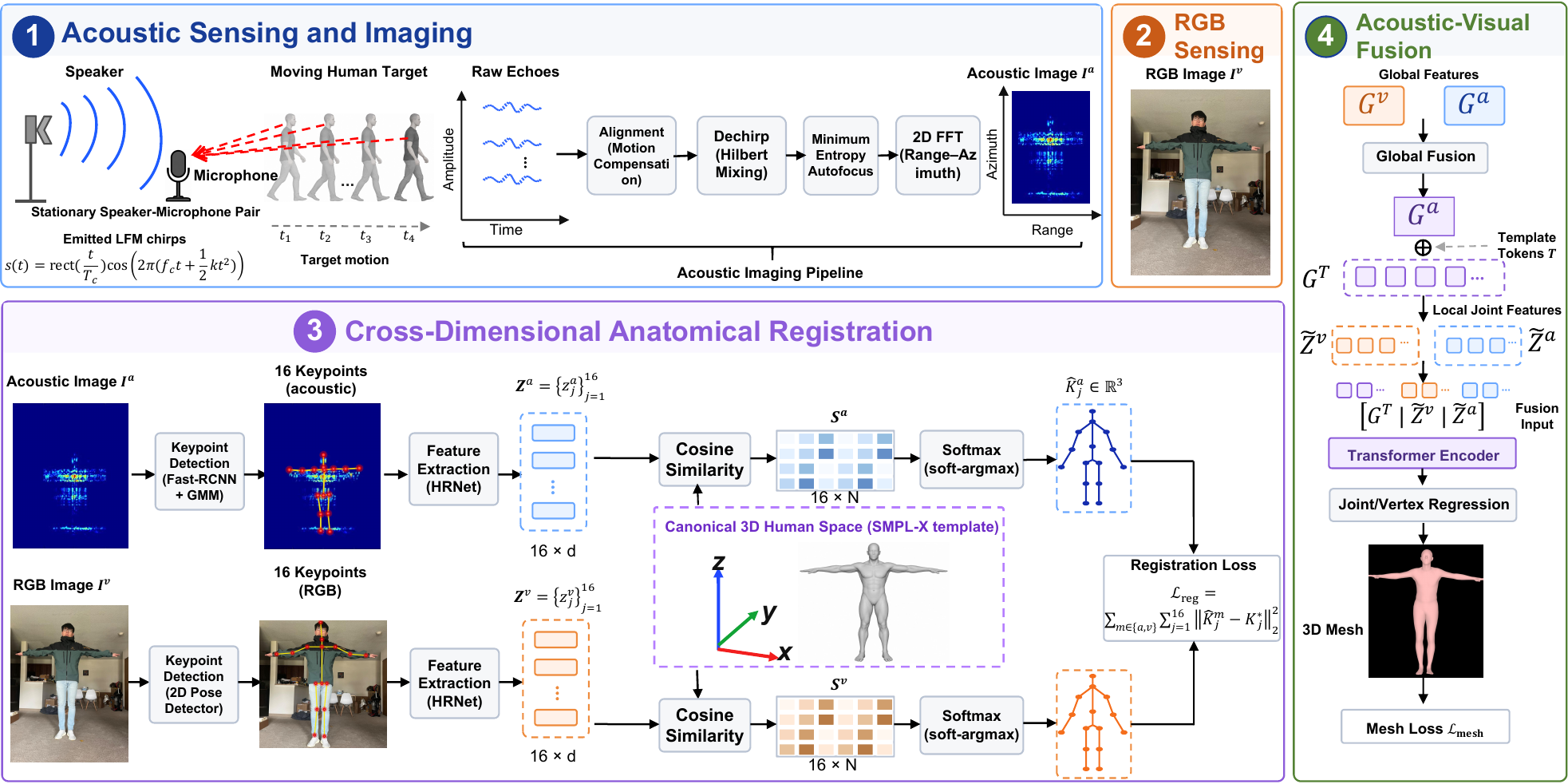}
                        \caption{Overview of the proposed acoustic--visual framework for 3D human mesh reconstruction, including acoustic imaging, cross-dimensional anatomical registration, and multimodal fusion.}
            \label{fig:framework}
\end{figure*}

\subsection{Cross-Dimensional Anatomical Registration}
\label{sec:registration}

Before cross-dimensional registration, we first localize the 2D anatomical keypoints in both modalities. For the RGB image, a standard 2D keypoint detector is used to obtain the $J=16$ anatomical joints. Due to the lower resolution and more diffuse boundaries of acoustic images, we adopt a two-stage localization strategy for the acoustic modality. Specifically, Fast R-CNN~\cite{girshick2015fast} is first used to detect the human region, followed by a Gaussian mixture model (GMM) to segment the acoustic responses into body-related regions. The joint locations are then estimated from the geometric centers and high-density regions of the corresponding segments (see Appendix~\ref{app:registration_details} for details), yielding
\begin{equation}
\mathcal{K}^{m}
=
\{k_j^m\}_{j=1}^{J},
\qquad
k_j^m\in\mathbb{R}^{2},
\quad m\in\{a,v\}.
\end{equation}

% For each modality $m\in\{a,v\}$, we first extract a feature map
% \begin{equation}
% F^{m}=E_m(I^{m}),
% \end{equation}
% where $E_m(\cdot)$ denotes the modality-specific feature encoder.

For each modality $m\in\{a,v\}$, we employ a modality-specific HRNet~\cite{sun2019deep} to extract a dense feature map:
\begin{equation}
F^m=\operatorname{HRNet}_m(I^m).
\end{equation}

For the $j$-th detected anatomical keypoint $k_j^m\in\mathcal K^m$, the corresponding local feature is sampled from $F^m$ as:
\begin{equation}
z_j^m
=
\operatorname{Sample}(F^m,k_j^m)
\in\mathbb R^d.
\label{eq:joint_feature}
\end{equation}

% We further define a canonical 3-D human space containing candidate body locations
% \begin{equation}
% \mathcal{C}=\{c_i\}_{i=1}^{N}, \qquad c_i\in\mathbb{R}^3,
% \end{equation}

We define a canonical 3D anatomical space based on the SMPL-X body template~\cite{pavlakos2019expressive} and uniformly sample a $20\times20\times20$ grid within its bounding volume, yielding $N=8000$ candidate locations
$\mathcal{C}=\{c_i\}_{i=1}^{N}$, where $c_i\in\mathbb{R}^3$. Here, $c_i$ denotes a canonical candidate location, while $K_j^*$ denotes the ground-truth 3D position of the $j$-th joint in the same canonical coordinate system. Each candidate location is embedded into the same feature space through an MLP,
\begin{equation}
h_i=g(c_i)
\in\mathbb{R}^{d}.
\end{equation}
We then measure the cosine similarity between the $j$-th 2D joint feature and the $i$-th candidate location:
\begin{equation}
s_{ji}^{m}
=
\frac{
(z_j^{m})^{\top} h_i
}{
\|z_j^{m}\|_2\,
\|h_i\|_2
}.
\label{eq:registration_similarity}
\end{equation}
which are then normalized using a softmax function,
\begin{equation}
w_{ji}^{m}
=
\frac{
\exp(s_{ji}^{m})
}{
\sum_{\ell=1}^{N}
\exp(s_{j\ell}^{m})
},
\end{equation}
The registered 3D location is then obtained as the
softmax-weighted expectation over the candidate locations:
\begin{equation}
\hat{K}_j^{m}
=
\sum_{i=1}^{N}
w_{ji}^{m} c_i.
\label{eq:registered_joint}
\end{equation}

%In this way, modality-specific 2D observations are anchored to a shared canonical 3D anatomical space, providing explicit structural guidance for subsequent acoustic--visual fusion.

%During training, the registration module is jointly optimized with the reconstruction network. 

%Given the ground-truth 3-D joint location $K_j^{*}$, we define the registration loss as:

%To evaluate the alignment between predicted 3D points and their ground-truth 3-D joint location $K_j^{*}$, we define the registration loss as:

To supervise the registration, we minimize the distance between the predicted and ground-truth 3D joint locations:
\begin{equation}
\mathcal{L}_{\mathrm{reg}}
=
\sum_{m\in\{a,v\}}
\sum_{j=1}^{J}
\left\|
\hat{K}_j^{m}
-
K_j^{*}
\right\|_2^2.
\label{eq:registration_loss}
\end{equation}

%This auxiliary supervision encourages both acoustic and visual features to preserve anatomically consistent 3-D structure before multimodal fusion.

\subsection{Acoustic--Visual Fusion and Mesh Reconstruction}
\label{sec:fusion}

Besides the joint-level local features extracted in
Sec.~\ref{sec:registration}, we obtain a global representation for each modality by aggregating its dense feature map:
\begin{equation}
G^m=\operatorname{Pool}(F^m),
\qquad m\in\{a,v\}.
\label{eq:global_feature}
\end{equation}
The global features capture the overall body structure and pose, whereas the joint-level features $Z^m=[z_1^m,\ldots,z_J^m]$ preserve fine-grained anatomical
information.

After cross-dimensional registration, each joint feature $z_j^m$ is associated with an estimated canonical 3D location $\hat K_j^m$. We concatenate them to construct the anatomy-aware representation:
\begin{equation}
\tilde z_j^m
=
\left[
z_j^m;\hat K_j^m
\right].
\end{equation}
We denote the resulting joint-level representations by
$\tilde Z^m=[\tilde z_1^m,\ldots,\tilde z_J^m]$.

We then adopt a two-stage fusion strategy. We first fuse the modality-specific global features $G^a$ and $G^v$
using a tiny Transformer module, and concatenate the resulting global representation $G$ with the canonical SMPL-X template $T$:
\begin{equation}
G = f_{\mathrm{fus}}(G^a,G^v), \qquad
G^T = \operatorname{cat}(G,T).
\end{equation}

% where $T$ is the embedding of the canonical SMPL-X body template~\cite{pavlakos2019expressive}..

% \begin{equation}
% Z
% =
% \left[
% G^{T},\,
% \tilde{Z}^a,\,
% \tilde{Z}^v
% \right].
% \label{eq:fusion_input}
% \end{equation}
% The resulting representation is processed by a Transformer encoder,
% \begin{equation}
% H
% =
% \operatorname{Transformer}(Z),
% \label{eq:fusion_transformer}
% \end{equation}

The template-aware global representation $G^T$ is then combined with the anatomy-aware joint representations $\tilde Z^a$ and $\tilde Z^v$ using a multi-head Transformer~\cite{vaswani2017attention}:
\begin{equation}
H =
f_{\mathrm{tr}}
\left(
G^T,\tilde Z^a,\tilde Z^v
\right).
\end{equation}
The output $H$ is subsequently used to regress 3D joint coordinates and coarse mesh vertices, followed by an MLP-based refinement to obtain the final 3D human mesh $\hat{\mathcal{V}}$.

During training, modality masking is applied to randomly suppress either the acoustic or visual input, preventing the network from over-relying on a single modality and improving robustness under degraded sensing conditions (see Appendix~\ref{app:modality_masking} for details). The overall training objective is:
\begin{equation}
\mathcal{L}
=
\mathcal{L}_{\mathrm{mesh}}
+
\lambda
\mathcal{L}_{\mathrm{reg}},
\label{eq:overall_loss}
\end{equation}
%where $\mathcal{L}_{\mathrm{mesh}}$ is the loss of 3-D mesh reconstruction, $\mathcal{L}_{\mathrm{reg}}$ is the loss of cross-dimensional anatomical registration (Eq.~(\ref{eq:registration_loss})). 
where $\mathcal{L}_{\mathrm{mesh}}
= \frac{1}{N_v}\sum_{i=1}^{N_v}
\|\hat{\mathbf v}_i-\mathbf v_i^*\|_1$
denotes the mesh reconstruction loss, with $N_v$ being the number of mesh vertices, and $\mathcal{L}_{\mathrm{reg}}$ is the anatomical registration loss defined in Eq.~(\ref{eq:registration_loss}). The complete SonicMesh framework is illustrated in Fig.~\ref{fig:framework}.

%defined in Sec.~\ref{sec:registration}.

\section{Experiments}

%\subsection{Dataset}\label{sec:dataset}

%Inspired by~\cite{chen2022mmbody}, we developed a comprehensive dataset to evaluate SonicMesh's performance. 

As no public dataset provides the acoustic--visual modalities required for our task, we construct a new dataset to evaluate SonicMesh. We collected data from 20 volunteers (10 males and 10 females) performing eight daily activities (standing, arm spreading, single- and double-arm raising, diagonal arm extension, overhead extension, downward stretching, and side arm raising) across four indoor environments: a conference room, laboratory, corridor, and living room. The dataset covers normal, poor-light, occluded, and NLOS conditions, each accounting for \(25\%\) of the collected data. We adopt a strictly subject-independent split with 14, 2, and 4 participants for training, validation, and testing, with no participant overlap across subsets and a balanced gender distribution in each split. All experiments are repeated 10 times with different seeds, and the results are averaged. Ground truth is captured using a calibrated camera system. Additional implementation and dataset details are provided in Appendix~\ref{app:implementation_details}.

%The dataset's design emphasizes diversity across multiple dimensions - from participant demographics and body types to movement patterns and environmental settings. This comprehensive coverage enables SonicMesh to learn generalizable features that transfer well to real-world applications, where varying conditions and user characteristics are common. We plan to open-source the dataset upon paper acceptance to support further research in this area.

\begin{table}[t]
\centering
\caption{Comparison with state-of-the-art methods. Results are reported as MPJPE/PVE (cm).}
\label{tab:mpjpe}
\resizebox{\columnwidth}{!}{
\begin{tabular}{lcccccc}
\toprule
Method & Signal & Normal & Poor-light & Occlusion & NLOS & Avg. \\
\midrule
mmMesh~\cite{xue2021mmmesh}
& mmWave & 6.53/8.12 & 6.47/7.95 & 7.06/8.23 & 6.95/8.15 & 6.80/8.12 \\

RF-Avatar~\cite{zhao2019through}
& RF & 7.05/8.92 & 7.13/9.03 & 7.07/9.15 & 7.26/9.21 & 7.13/9.07 \\

WiMesh~\cite{wang2022wi}
& RF & 6.41/7.10 & 6.43/7.09 & 6.34/7.21 & 6.57/7.43 & 6.50/7.21 \\

ImmFusion~\cite{chen2023immfusion}
& mmWave & 6.50/7.70 & 6.62/7.79 & 6.80/8.13 & 7.10/8.09 & 6.79/7.93 \\

Graphormer~\cite{lin2021mesh}
& RGB & 5.12/7.31 & 46.32/54.47 & 28.70/34.55 & 59.24/63.53 & 34.84/39.97 \\

Deformer~\cite{yoshiyasu2023deformable}
& RGB & 4.48/6.14 & 35.47/44.03 & 27.50/31.23 & 49.76/54.15 & 29.30/33.89 \\

% SonicMesh-RGB
% & RGB & 5.31/6.61 & 30.53/39.67 & 15.70/23.55 & 42.43/51.12 & 23.49/30.23 \\

% SonicMesh-Acoustic
% & Acoustic & 6.43/7.05 & 4.97/6.19 & 6.27/7.09 & 6.93/8.75 & 6.15/7.27 \\

\textbf{SonicMesh}
& Acoustic+RGB & \textbf{5.31/6.43} & \textbf{5.39/6.54}
& \textbf{5.43/6.57} & \textbf{6.92/8.72}
& \textbf{5.81/7.10} \\
\bottomrule
\end{tabular}}
\end{table}

\begin{figure}[ht]
    \centering
    \includegraphics[width=0.9\linewidth]{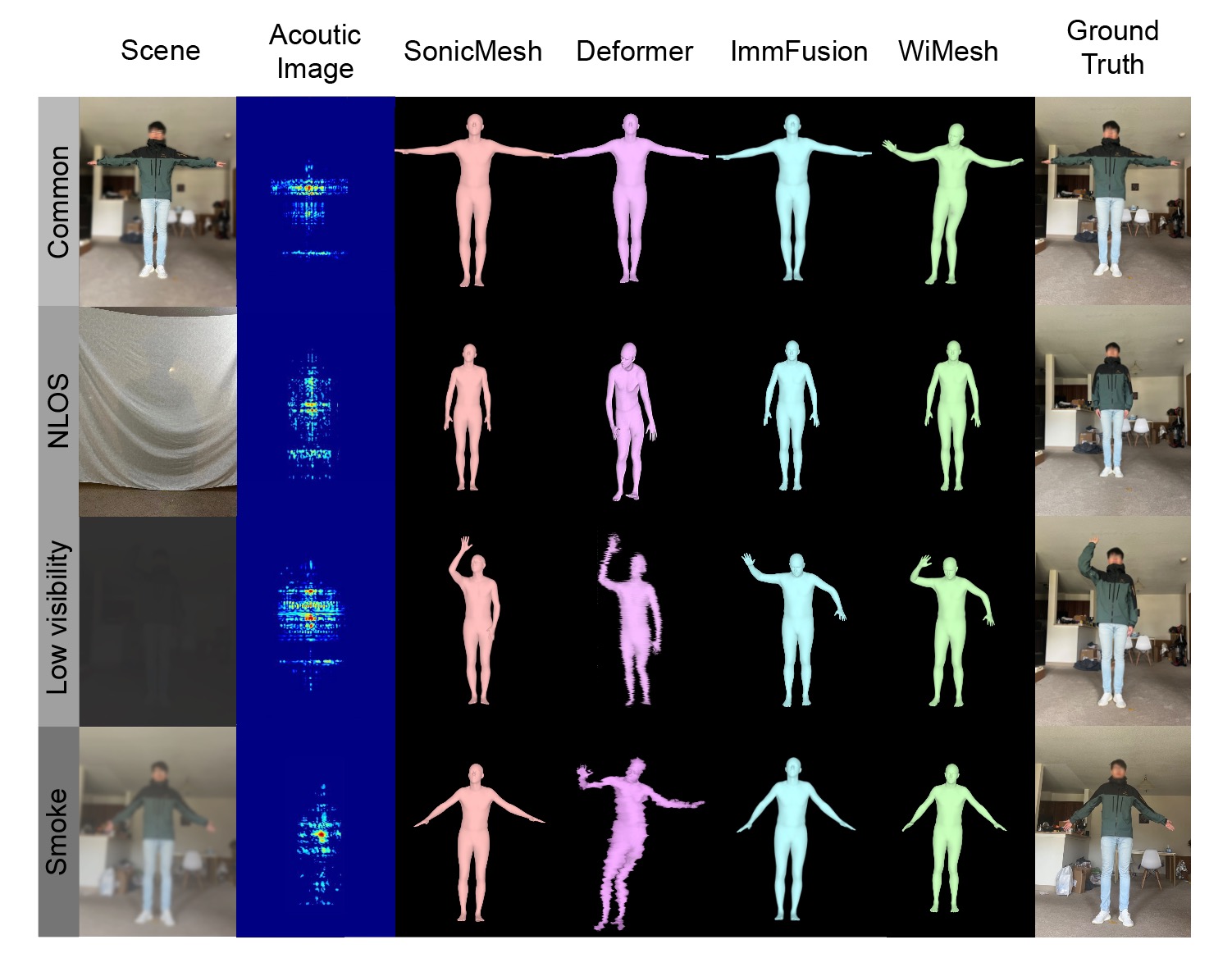} 
    \caption{Qualitative results (compared with Deformer, ImmFusion, and WiMesh in common, NLOS, low visibility, and smoke scenes).}
    \label{fig:result}
\end{figure}

We evaluate SonicMesh using Per Vertex Error (PVE) and Mean Per Joint Position Error (MPJPE). Lower values indicate better accuracy.
For RF/mmWave baselines, we adapt only their modality-dependent input and output layers to our acoustic input and mesh output, while preserving the core architectures.
As shown in Table~\ref{tab:mpjpe} and Fig.~\ref{fig:result}, SonicMesh produces accurate 3-D human meshes across normal, poor-light, occluded, and NLOS conditions. While RGB-based methods such as Graphormer~\cite{lin2021mesh} and
Deformer~\cite{yoshiyasu2023deformable} perform well under normal visibility, their accuracy degrades substantially under adverse visual conditions. In contrast, SonicMesh maintains stable reconstruction performance and consistently outperforms representative mmWave- and
RF-based approaches, including ImmFusion~\cite{chen2023immfusion} and WiMesh~\cite{wang2022wi}, in challenging environments. Beyond accuracy, SonicMesh also relies only on commodity microphones, speakers, and RGB cameras, whereas mmWave-~\cite{chen2023immfusion,xue2021mmmesh} and
RF-based methods~\cite{wang2022wi,zhao2019through} typically require dedicated radar hardware or specific WiFi antenna configurations.

% \subsection{Ablation and Sensitivity Analysis}
% \label{sec:benchmark}

Table~\ref{tab:ablation} evaluates the contributions of both sensing modalities and key model components. The acoustic-only variant substantially outperforms the RGB-only variant, while combining both modalities further improves the average MPJPE/PVE to $5.81/7.10$~cm, demonstrating their complementarity. Removing the registration module degrades MPJPE/PVE to $7.87/9.39$~cm.
Replacing the proposed canonical registration with direct MLP-based 3D joint regression also leads to inferior performance. 
Further removing the acoustic localization components (Fast R-CNN and GMM) increases them to $10.58/13.63$~cm, confirming the effectiveness of the proposed components.

% \begin{table}
% \centering
% \caption{Ablation study}
% \resizebox{0.96\linewidth}{!}{%
% \begin{tblr}{
%   width = \linewidth,
%   colspec = {Q[537]Q[125]Q[194]},
%   column{2} = {c},
%   column{3} = {c},
%   hline{1-2,5} = {-}{},
% }
%                        & {PVE\\(cm)} & {MPJPE\\(cm)} \\
% SonicMesh              & 7.10           & 5.81             \\
% w/o Registration Module (RM) & 9.39           & 7.87             \\
% w/o Fast-CNN + GMM + RM      & 13.63           & 10.58             
% \end{tblr}
% }
% \label{comparison_module}
% \end{table}

\begin{table}[t]
\centering
\caption{Ablation study of sensing modalities and model components.}
\label{tab:ablation}
\resizebox{0.98\linewidth}{!}{
\begin{tabular}{lcc}
\toprule
Variant & PVE (cm) & MPJPE (cm) \\
\midrule
SonicMesh-RGB only & 30.23 & 23.49 \\
SonicMesh-Acoustic only & 7.27 & 6.15 \\
\midrule
w/o Registration Module (RM) & 9.39 & 7.87 \\
w/ Direct 3D Regression & 7.86 & 6.47 \\
w/o Fast R-CNN + GMM + RM & 13.63 & 10.58 \\
\midrule
\textbf{SonicMesh} & \textbf{7.10} & \textbf{5.81} \\
\bottomrule
\end{tabular}}
\end{table}

\begin{figure}[ht]
    \centering
    \includegraphics[width=0.8\linewidth]{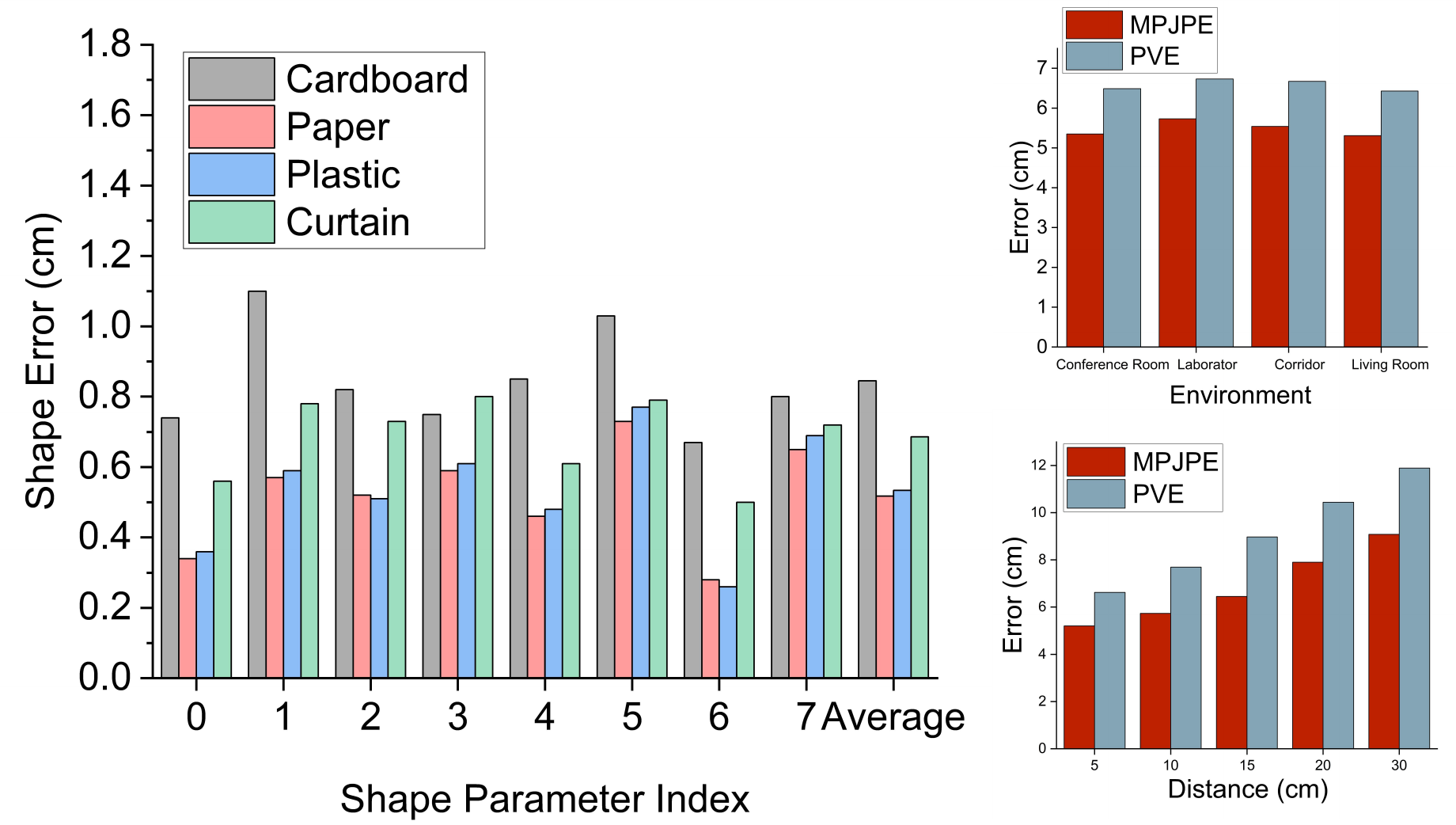} 
    \caption{Robustness under different occlusions (left), environments (top right), and speaker--microphone distances (bottom right).}
    \label{fig:robustness}
\end{figure}

We finally evaluate the robustness of SonicMesh under different sensing conditions, as shown in Fig.~\ref{fig:robustness}. For the distance
analysis, tests are conducted with speaker--microphone separations of $5$~cm, $10$~cm, $15$~cm, $20$~cm, and $30$~cm, showing that reconstruction accuracy gradually degrades as the distance increases. We also evaluate four indoor environments with different room sizes and multipath characteristics: a living room ($4$~m $\times$ $3$~m), a laboratory ($20$~m $\times$ $10$~m), a corridor ($3$~m $\times$ $20$~m), and a conference room ($5$~m $\times$ $7$~m). SonicMesh maintains relatively stable performance across these environments. In addition, experiments with different occlusion materials (e.g., plastic and paper) show that the method remains robust in most cases, although strongly reflective or absorbing materials result in larger reconstruction errors.

\section{Conclusion}
In this paper, we introduce SonicMesh, the first method to combine acoustic and RGB images for 3D human mesh reconstruction without handcraft features. Our approach addresses fundamental limitations of traditional camera-based methods through two key technical innovations. First, we incorporate a registration module that aligns pixel feature embeddings from 2D images with 3D canonical feature embeddings, effectively addressing the depth ambiguity inherent in 2D-to-3D reconstruction. Second, we enhance the HRNet architecture to better process low-resolution acoustic images, enabling effective feature extraction despite the quality limitations of acoustic signals. The system maintains robust performance across challenging scenarios where traditional methods fail, including non-line-of-sight (NLOS) conditions, poor lighting environments, and occluded scenes. By effectively combining the complementary strengths of acoustic and RGB modalities, SonicMesh represents a significant advance in reliable 3D human mesh reconstruction.

%\vfill
%\pagebreak

\appendix

\section{Additional Acoustic Imaging Details}
\label{app:acoustic_details}

\begin{figure}[ht]
    \centering
    \begin{subfigure}[b]{0.22\textwidth}
        \centering
        \includegraphics[height=0.95in]{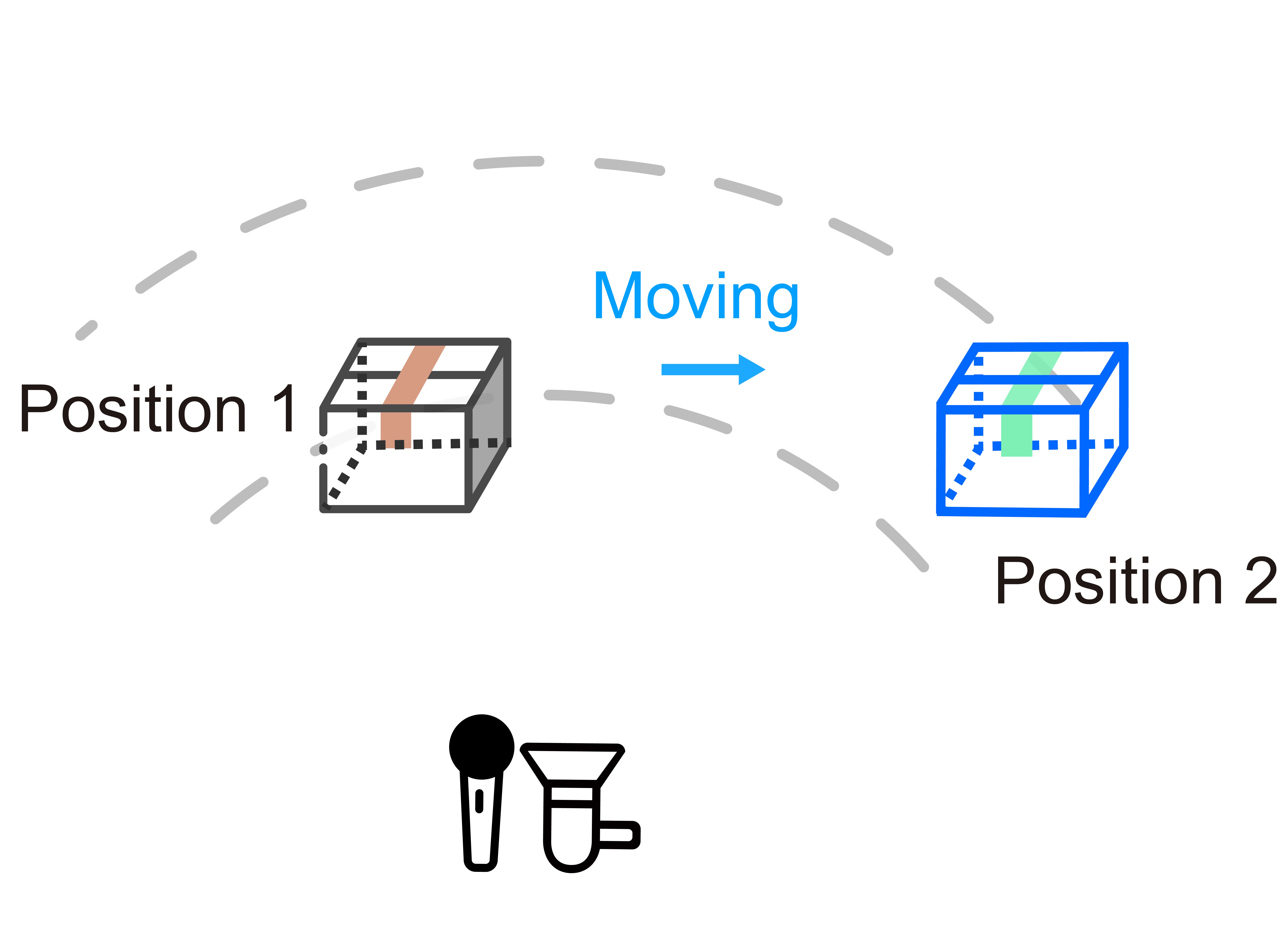}
        \caption{Target's actual motion.}
        \label{fig:subfig:actual}
    \end{subfigure}
    \hfill
    \begin{subfigure}[b]{0.22\textwidth}
        \centering
        \includegraphics[height=0.85in]{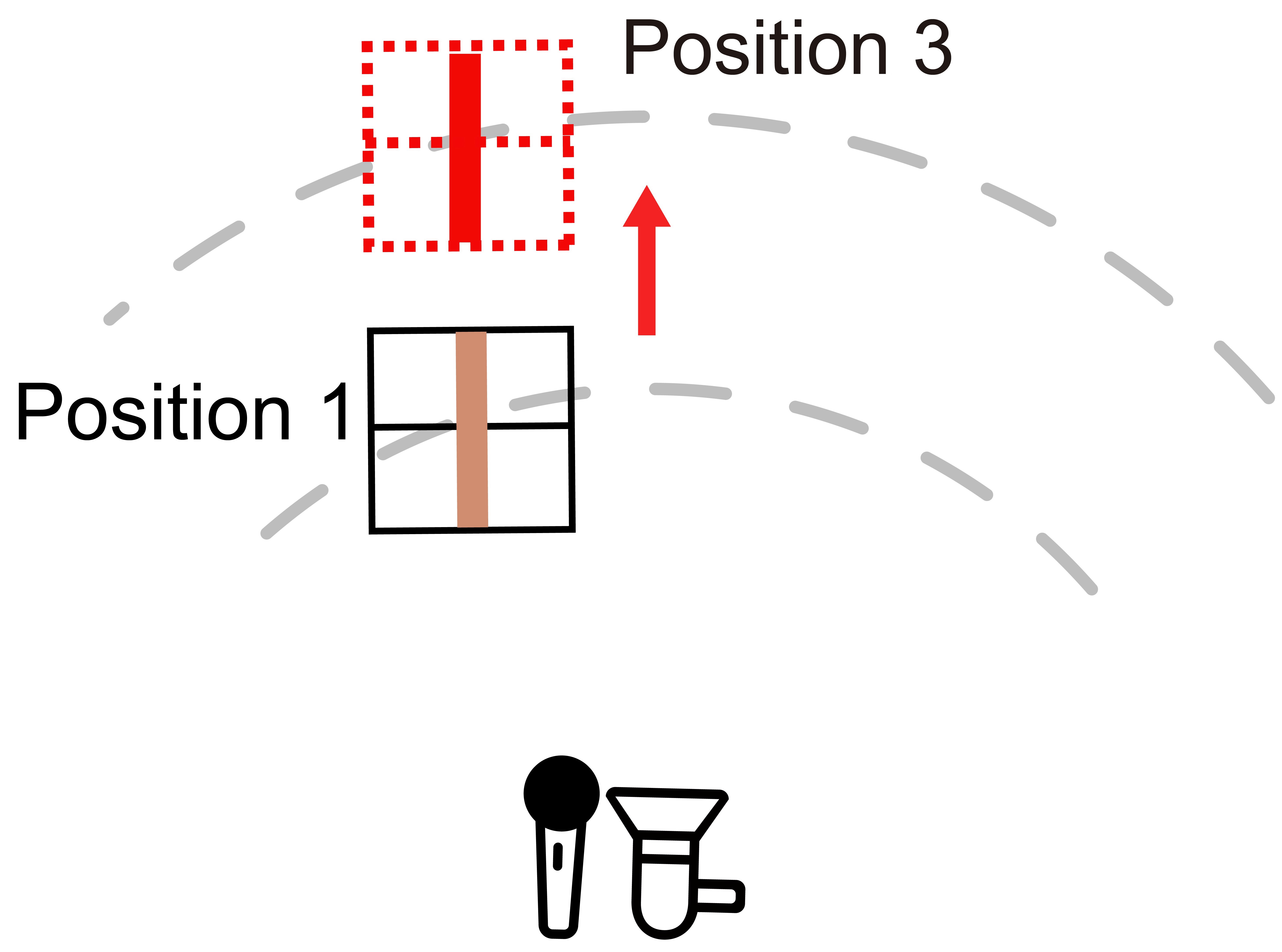}
        \caption{Step 1: translation motion.}
        \label{fig:subfig:1step}
    \end{subfigure}
    \hfill
    \begin{subfigure}[b]{0.22\textwidth}
        \centering
        \includegraphics[height=0.9in]{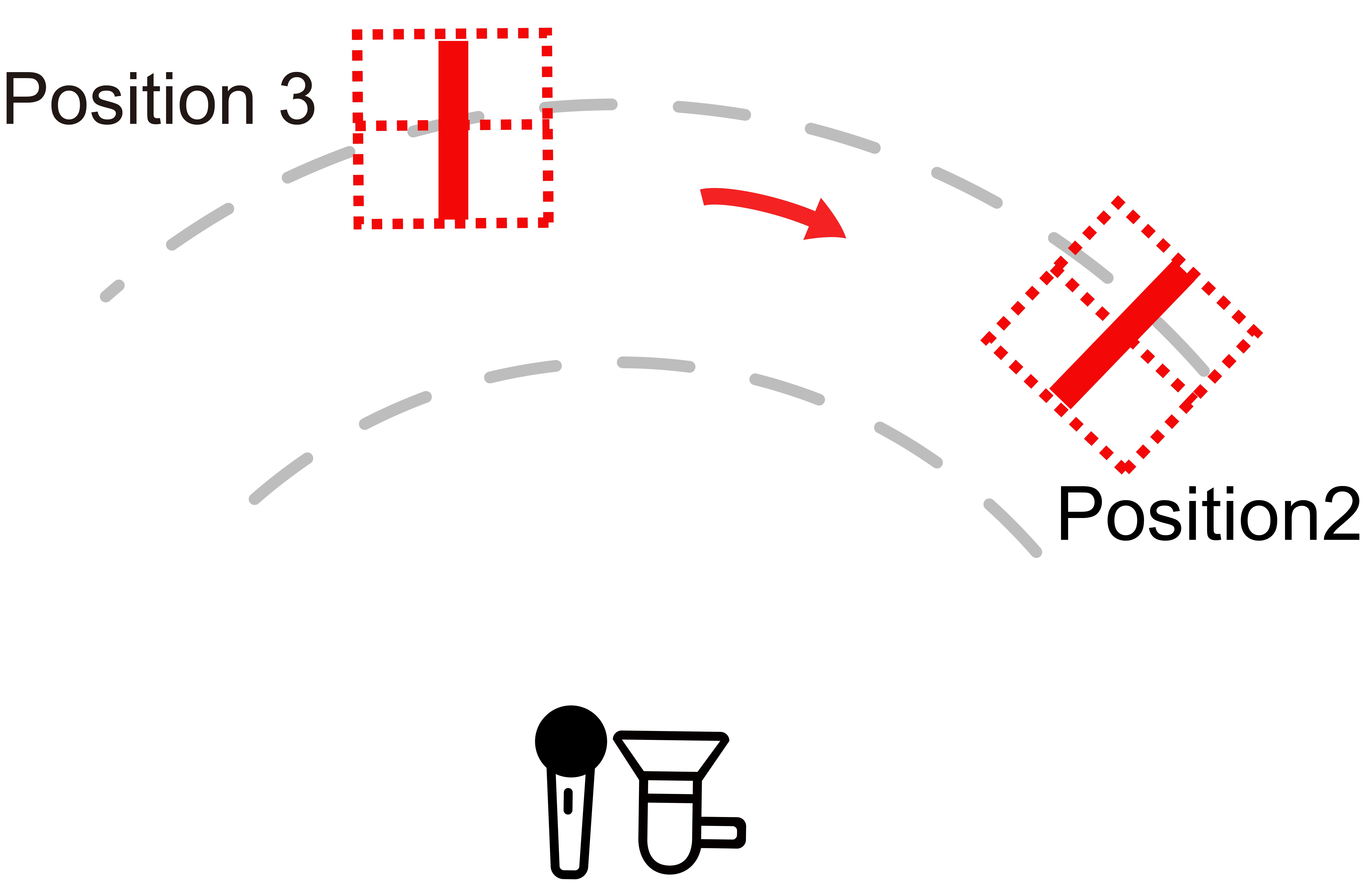}
        \caption{Step 2: circular motion.}
        \label{fig:subfig:2step}
    \end{subfigure}
    \hfill
    \begin{subfigure}[b]{0.22\textwidth}
        \centering
        \includegraphics[height=0.85in]{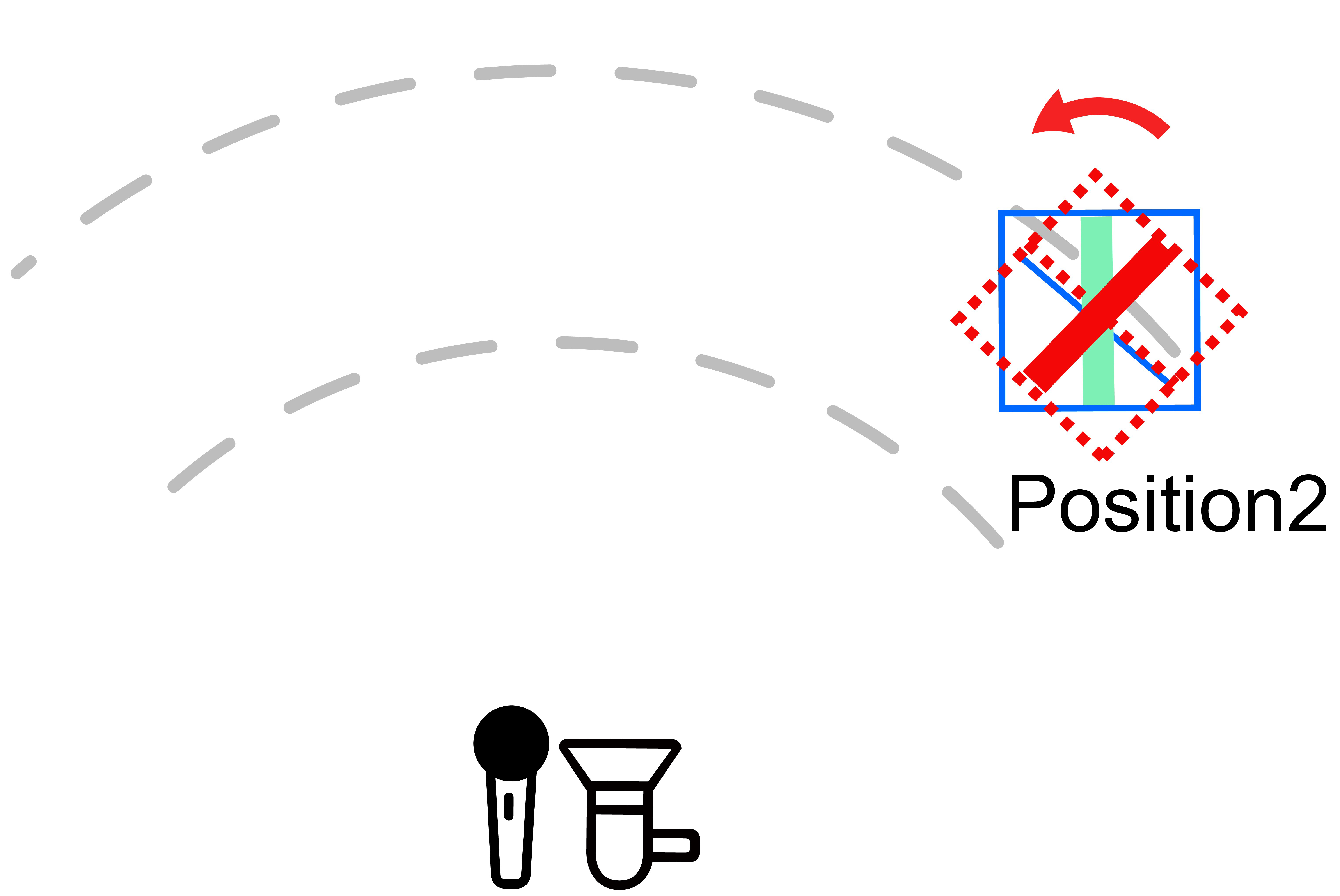}
        \caption{Step 3: rotation motion.}
        \label{fig:subfig:3step}
    \end{subfigure}
    \caption{The target's actual motion can be decomposed into three steps: (a) translation motion, (b) circular motion, and (c) rotation motion.}
    \label{fig:target_motion}
\end{figure}

\subsection{The principle of ISAR} \label{sec:principle_ISAR}

Inverse Synthetic Aperture Radar (ISAR) is commonly employed in military applications to identify the shapes of moving targets, such as ships. Unlike motion tracking systems that consider the target as a single point, ISAR regards the target as an array of scatter points. ISAR imaging converts the target’s translational motion into rotational motion, allowing for the capture of reflected signals from various perspectives around the target.

As shown in Fig.\ref{fig:subfig:actual}, consider a target moving from Position 1 to Position 2. This movement is illustrated in three stages in Fig.\ref{fig:subfig:1step}-\ref{fig:subfig:3step}. First, the target moves radially from Position 1 to Position 3 (Fig.\ref{fig:subfig:1step}), resulting in additional propagation delays in the signals reflected from each scatter point. These delays introduce image distortion, which is corrected using our echo alignment technique. In the second stage, the target moves along a circular path centered on the imaging device, from Position 3 to Position 2 (Fig.\ref{fig:subfig:2step}), without affecting propagation delays or the final image outcome. In the final stage, the target rotates around its geometric center by a specific angle (Fig.~\ref{fig:subfig:3step}), aligning with the azimuth angle traced during its actual movement.

\subsection{The Basics of Chirp Signal}
\label{sec:chirp_basic}
\begin{figure}[ht]
    \centering
    \begin{subfigure}[b]{0.4\textwidth}
        \centering
        \includegraphics[width=1.3in]{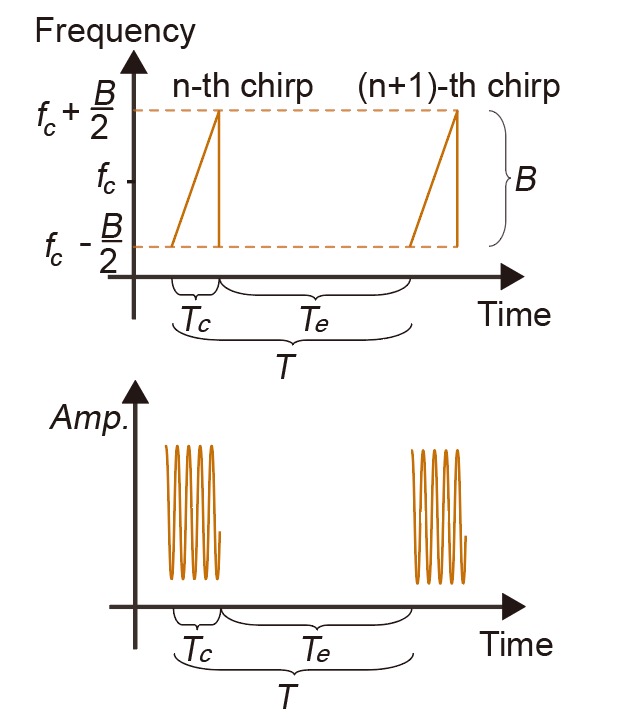}
        \caption{The chirp signal.}
        \label{subfig:chirp}
    \end{subfigure}
    \hfill
    \begin{subfigure}[b]{0.5\textwidth}
        \centering
        \includegraphics[width=1.8in]{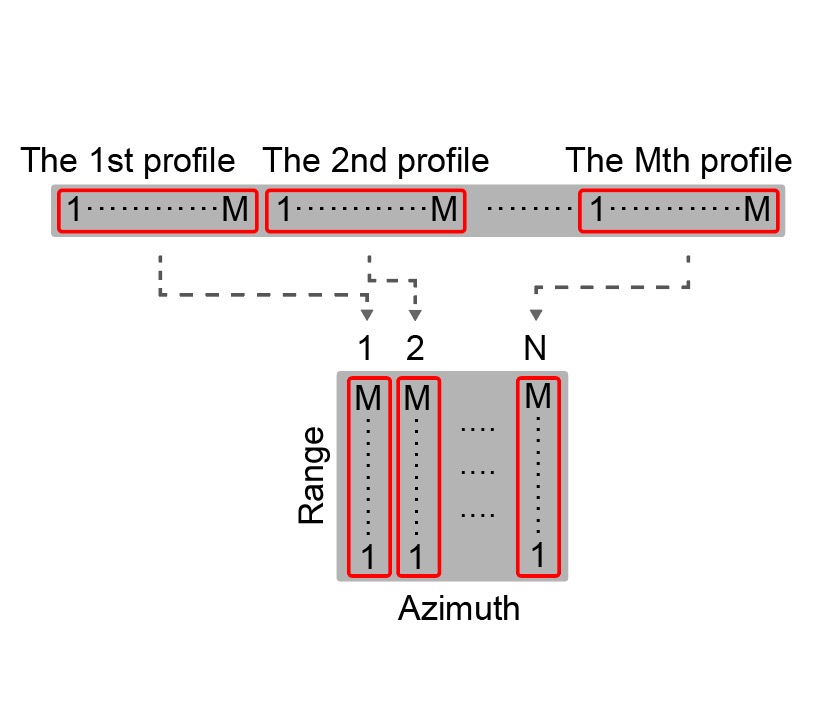}
        \caption{The profile matrix.}
        \label{fig:subfig:echo_matrix}
    \end{subfigure}
    \caption{The signal illustration for imaging.}
    \label{fig: signals}
\end{figure}

A speaker emits a series of chirp signals with frequencies that linearly sweep over time, as shown in Fig.~\ref{subfig:chirp}. The frequency of each chirp ranges from \(f_c - \frac{B}{2}\) to \(f_c + \frac{B}{2}\), where \(f_c\) is the carrier frequency and \(B\) is the bandwidth.  Each chirp lasts for a duration \(T_c\), followed by a pause, \(T_e\), to allow all echoes from one chirp to return before the next chirp is emitted. Each chirp can be mathematically expressed as
\begin{equation}
s(t) =  \mathrm{rect}\left[\frac{t}{T_c}\right]\cos\Big(2\pi \big(f_c t+\frac{k}{2}t^2\big)\Big),
\label{chirp_signal}
\end{equation}
where $k = \frac{B}{T_c}$ is the slope of the chirp. The term $\mathrm{rect}[\cdot]$ refers to the rectangle window function, which is defined as
\begin{equation}
\mathrm{rect}[\cdot] = \left\{ \begin{array}{ll}
1 & |\cdot| \leq 0.5 \\
0 & |\cdot| \geq 0.5. \\
\end{array} \right.
\label{rect}
\end{equation}
The total duration of each chirp cycle, including the pause, is $T = T_c + T_e$. The transmitted chirp signal is reflected by the target and received by a microphone, creating an echo. The time delay~($\Delta \tau$) of the echo is typically calculated by correlating the transmitted chirp signal with the received echo~\cite{mahafza2022radar}. Using this time delay, the distance $D$ etween the target and the device can be determined by:
\begin{equation}
    D = \frac{\Delta \tau \times v_s}{2}, 
    \label{new_distance}
\end{equation}
where $v_s$ is the sound propagation speed in the air.

\subsection{Signal Model for Imaging}
The received echo signals are organized into a profile matrix for imaging, as illustrated in Fig.~\ref{fig:subfig:echo_matrix}. This matrix arranges \(M\) samples per column with \(N\) consecutive collected profiles to construct the image. Then we apply the Hilbert transform to the echo signals, followed by dechirping each column to extract the intermediate frequency (IF) signal matrix. Subsequently, a 2-D Fast Fourier Transform (FFT) is performed on the IF signal matrix along both the range and azimuth dimensions. This step effectively maps each scatter point in physical space to a corresponding pixel in the image space, \(I\), where a scatter point \(p\) is identified by a peak at position \((p_r, p_a)\) in the 2-D image space, which can be represented as
\begin{equation}
    I(i_a, i_r) = A \sinc\left[\frac{2kT}{v_s}(i_r- p_r)\right]\times\sinc\left[\frac{2 \theta}{\lambda}(i_a - p_a)\right],
\label{point}
\end{equation}
where $A$ is a constant related to the amplitude and chirp duration, $v_s$ is the sound speed, $\lambda$ is the wavelength, $\theta$ is the angle that the target rotates during the imaging process, $i_r$ and $i_a$ are the indexes along the range dimension and azimuth dimension, respectively.

\section{Additional Registration Details}\label{app:registration_details}

\subsection{Definition of the 16 Human Joints}
\begin{figure}
   % [Update the figure here!]
    \centering
    \includegraphics[width= 3.5in]{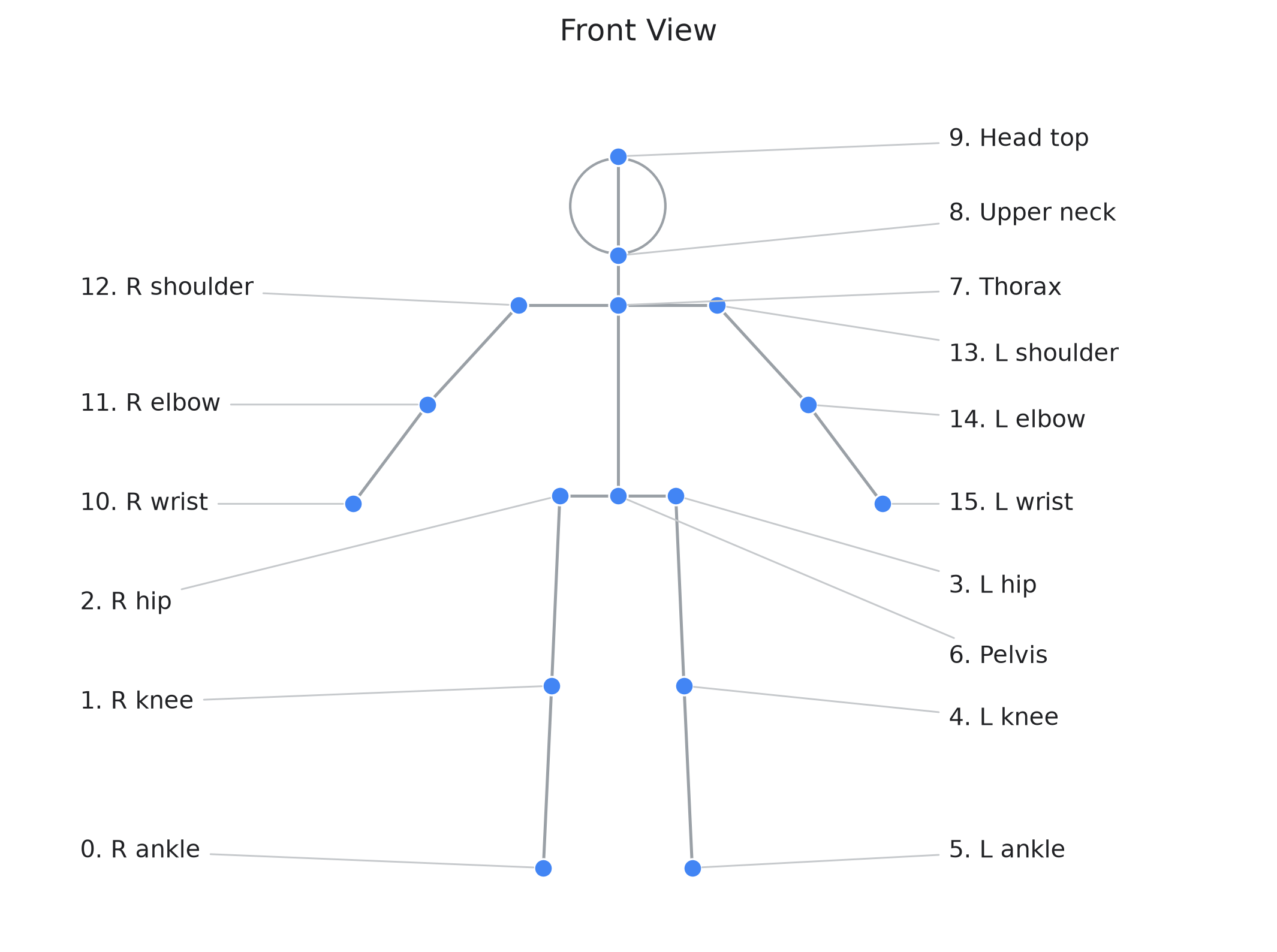}
    \caption{The 16 critical joints used in our registration module. The front view shows the primary joints, including the head, shoulders, elbows, wrists, hips, knees, and ankles.}
    \label{fig:joints}
\end{figure}

Fig.~\ref{fig:joints} shows the 16 human joints we considered.

\subsection{Why Soft-Argmax}

Soft-argmax provides a smooth and differentiable alternative to hard argmax for mapping a 2D joint feature to the canonical 3D space. For each modality $m\in\{a,v\}$, we compute the cosine similarity between the joint feature $z_j^m$ and the embedding $h_i=g(c_i)$ of each canonical candidate location $c_i\in\mathcal{C}$. The similarities are normalized by softmax to obtain correspondence weights $w_{ji}^m$ over the $N$ candidate
locations.

Rather than selecting only the candidate with the largest similarity, soft-argmax computes the registered 3-D location as the softmax-weighted expectation
\begin{equation}
    \hat K_j^m = \sum_{i=1}^{N} w_{ji}^m c_i .
\end{equation}
This formulation is fully differentiable, allowing the registration loss to propagate gradients through the correspondence weights and the modality-specific feature extractors. It also avoids the discontinuity of hard nearest-candidate selection.

\subsection{Acoustic Joint Localization}

First, we employ Fast-CNN~\cite{girshick2015fast} to detect the region of humans in acoustic images. Fast-CNN uses a bounding-box technique to efficiently scan acoustic images, identifying regions where shape and intensity patterns suggest human presence. This initial detection stage provides coarse spatial localization of human figures within the acoustic image. 

Second, to achieve robust pose estimation in human regions, we employ a Gaussian Mixture Model (GMM). GMM is particularly well-suited for this task because it models the underlying uncertainty and noise commonly found in acoustic data through a stochastic framework. 
Unlike deterministic clustering methods, GMM provides more stochasticity, allowing it to handle ambiguous boundaries between body parts gracefully. 
Furthermore, it naturally captures the spatial clustering of anatomical structures and implicitly encodes the relative positions and orientations of different segments, making it an effective tool for refining human body pose estimation.

We then locate 16 key joint points, each constrained to its corresponding segmented region as detected by the GMM. 
To locate each joint, we calculate the geometric center of each segment, $\bar{x} \triangleq \sum_{i=1}^{N}x_i / N$, where $x_i$ are points within a segment and $N$ is the total number of points in the segment, and analyze areas of maximal density as potential joint locations. This selection focuses on intersections and peripheries of these segments, minimizing the Euclidean distance $d(\mathbf{p}, \mathbf{q}) \triangleq \sqrt{\sum_{i=1}^{N}(q_i - p_i)^2}$ between centroids $\mathbf{p}$ and $\mathbf{q}$ of adjacent segments, as well as density measurements to identify clusters of data points within each segment.

\begin{figure}[ht]
    \centering
    \begin{subfigure}[b]{0.22\textwidth}
        \centering
        \includegraphics[width=\linewidth]{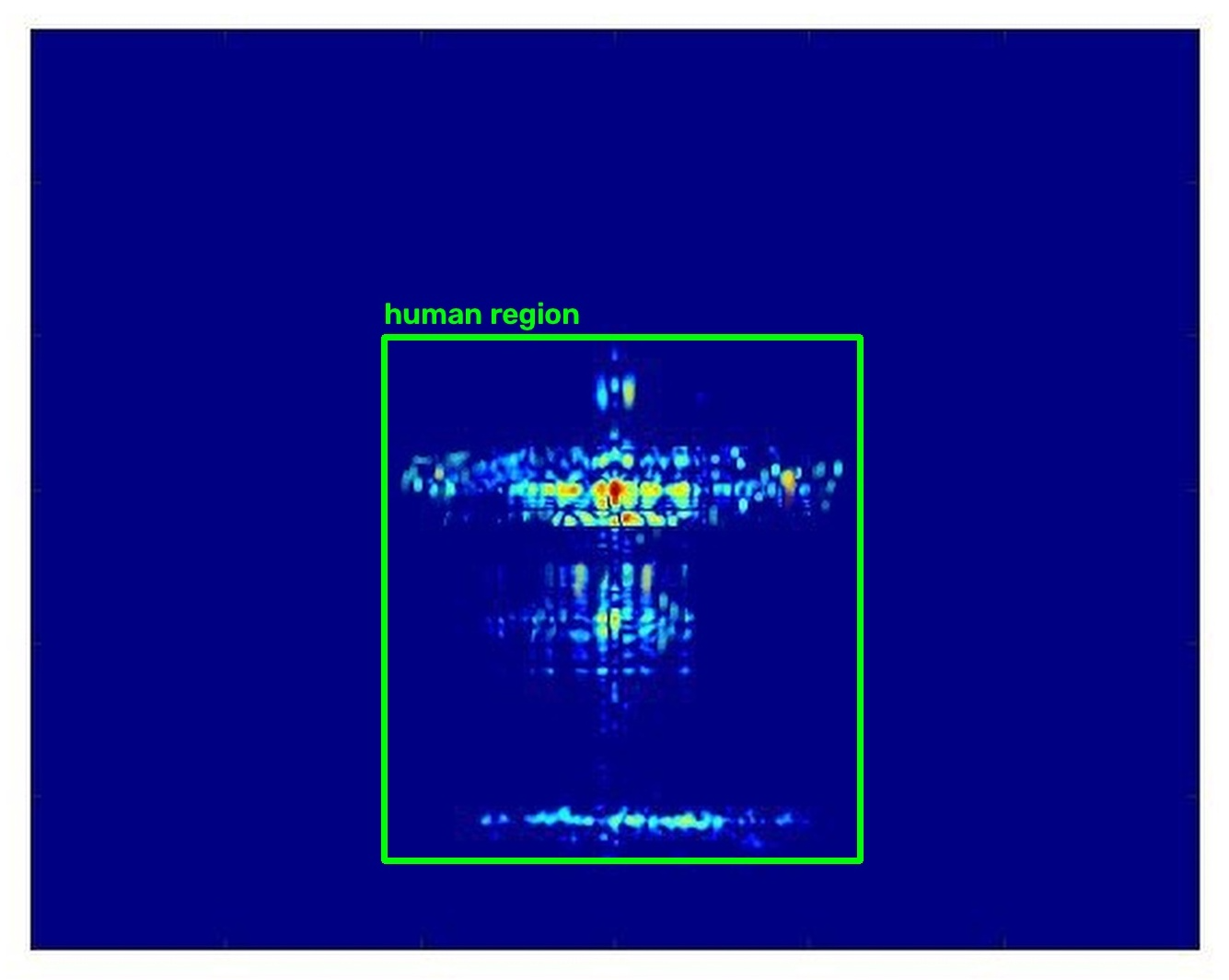}
        \caption{Detected human region.}
    \end{subfigure}
    \hfill
    \begin{subfigure}[b]{0.22\textwidth}
        \centering
        \includegraphics[width=\linewidth]{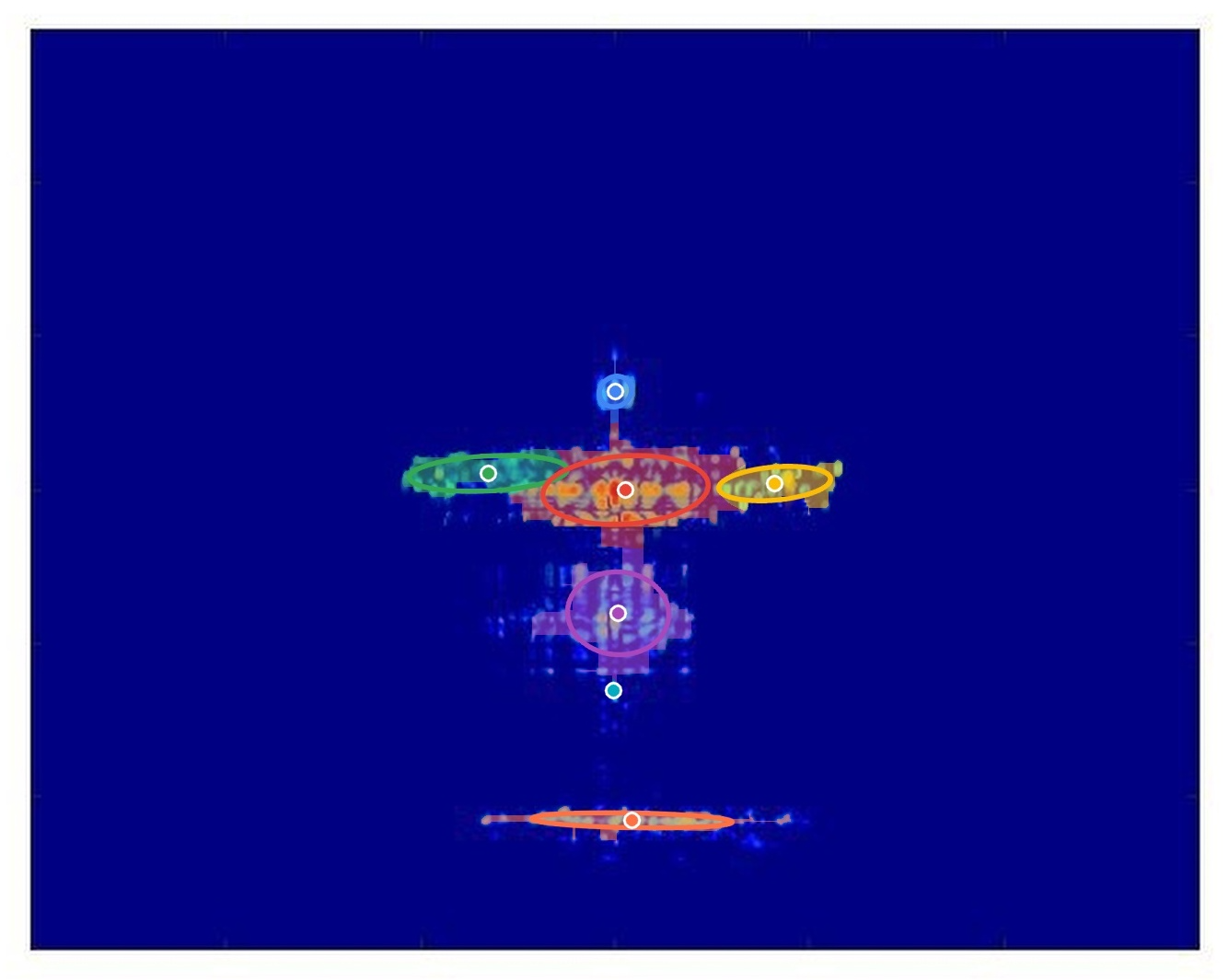}
        \caption{Body-region segmentation.}
    \end{subfigure}
    \hfill
    \begin{subfigure}[b]{0.22\textwidth}
        \centering
        \includegraphics[width=\linewidth]{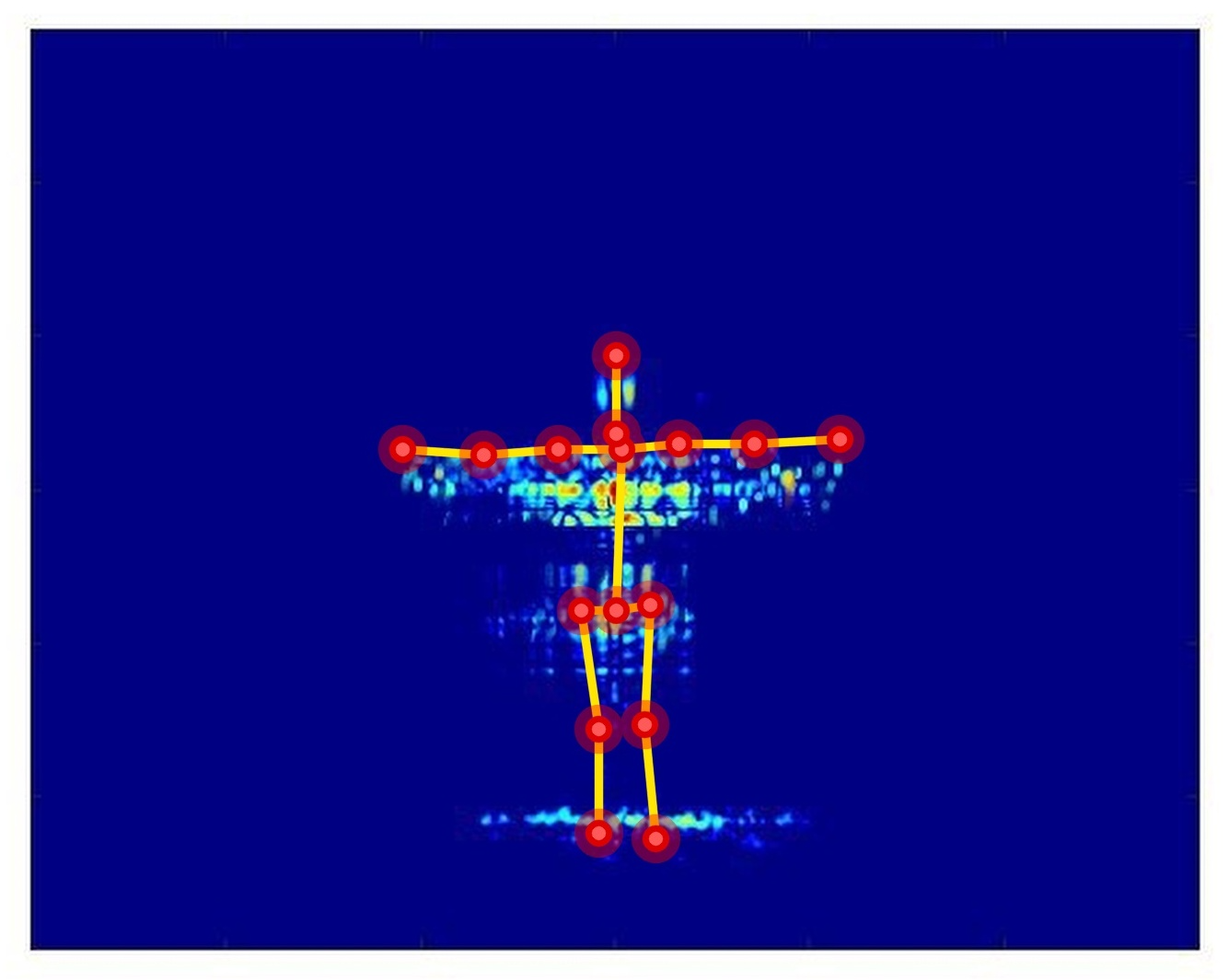}
        \caption{Estimated joint locations.}
    \end{subfigure}
    \caption{Illustration of acoustic joint localization:
    (a) human-region detection, (b) body-region segmentation, and (c) joint localization.}
    \label{fig:acoustic_localization}
\end{figure}

See Fig.~\ref{fig:acoustic_localization} for an illustrative example on acoustic joint localization.

\label{app:image_gen_by_aco}

\section{Modality Masking and Global--Local Fusion}\label{app:modality_masking}

Reliable 3D human mesh reconstruction requires consistent performance across diverse environments. While RGB and acoustic signals offer complementary strengths - RGB providing detailed features in good lighting and acoustic signals excelling in low visibility - naively combining these modalities creates a critical challenge. During training, models naturally gravitate toward the higher-quality RGB features in well-lit conditions~\cite{bijelic2020seeing, chen2023immfusion}. This bias manifests in two ways: the model fails to fully leverage acoustic signals when they would be most valuable (such as in poor lighting or occlusions), and simultaneously becomes overly reliant on RGB data quality, making the system vulnerable to environmental variations.

Understanding this fundamental challenge motivates our key technical contribution: strategies to enforce balanced learning from both modalities. By preventing over-reliance on either modality during training, we can ensure the system maintains robust performance regardless of environmental conditions, effectively utilizing whichever signal provides the most reliable information in any given situation.

To ensure balanced learning across modalities, we implement a two-pronged approach. First, we employ modality masking, which randomly occludes either RGB or acoustic inputs during training. This forces the model to learn robust features from each modality independently, preventing over-reliance on any single input source. Second, we incorporate a Global Integrated Module (GIM), inspired by~\cite{chen2023immfusion}, which dynamically weights the contribution of each modality based on input quality. While modality masking ensures independent feature learning, GIM optimizes the fusion of these features by adaptively combining information according to the reliability of each source.
Full details of the GIM architecture and training process can be found in \cite{chen2023immfusion}.

\section{Implementation Details}\label{app:implementation_details}

\subsection{Training Setup}
All models are implemented in Pytorch and trained on an \texttt{NVIDIA GeForce RTX 4090}. 
Each network is trained from scratch using the Adam optimizer with an initial learning rate of 0.2. The learning rate decays to 90\% of its previous value every 10 epochs, with a lower bound of 0.005.

\subsection{Acoustic device and configuration}
We implement our proof-of-concept system using a commodity device: a Bela~\cite{moro2016making} platform.  Signal processing and analysis are carried out using MATLAB.

\textbf{Bela Platform:} The Bela~\cite{moro2016making} platform is favored in acoustic sensing research for its versatility in supporting various microphone and speaker configurations. We equipped the Bela with a PUI Audio AS03104MR-N50-R speaker for signal transmission and a SparkFun BOB-18011 MEMS microphone for signal reception.

\textbf{Acoustic Signals:} To avoid interference from environmental noise, which typically falls below $16~kHz$~\cite{liang2024sondar}, we configure our system to sweep frequencies from $18~kHz$ to $22~kHz$. The duration of each chirp is set as $1~ms$, and the empty duration is set as $5~ms$.

\subsection{Dataset Details}
\begin{figure}[ht]
    \centering
    \begin{center}
        \begin{subfigure}[b]{0.2\textwidth}
            \centering
            \includegraphics[width=1.5in]{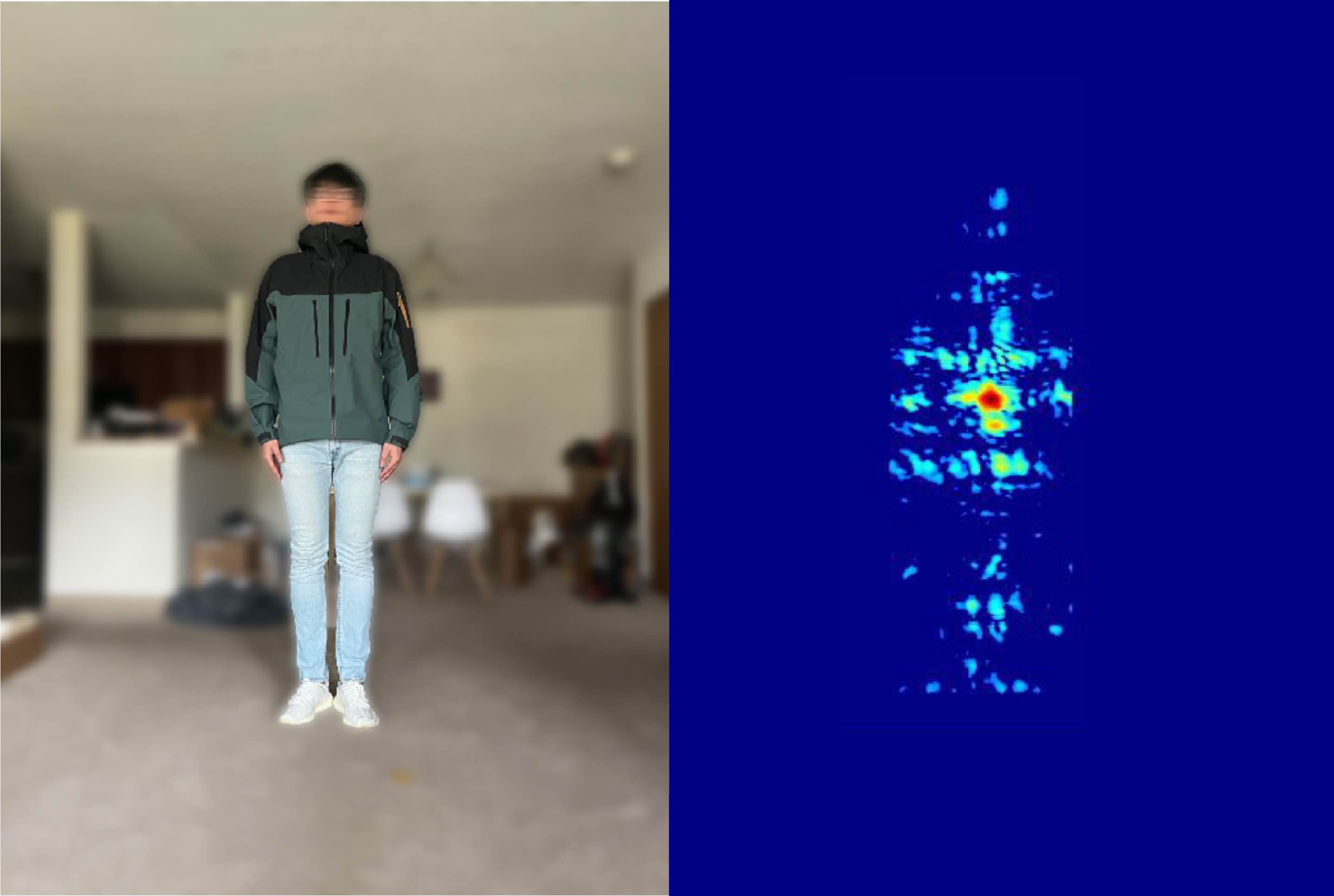}
            \caption{Standing.}
            \label{subfig:standing}
        \end{subfigure}
        \hfill
        \begin{subfigure}[b]{0.2\textwidth}
            \centering
            \includegraphics[width=1.5in]{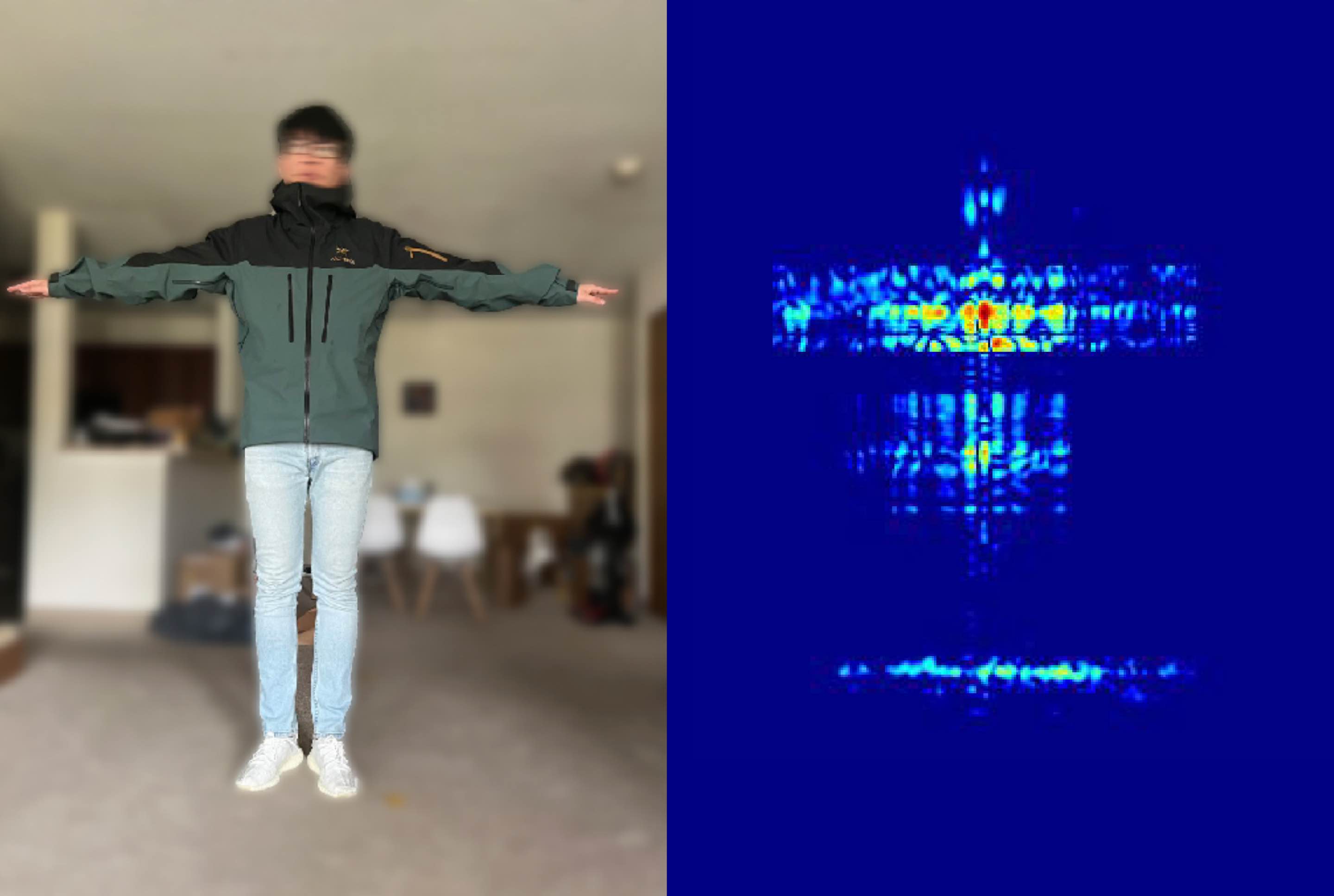}
            \caption{Arms spreading.}
            \label{fig:subfig:arm-spread}
        \end{subfigure}
        \hfill
        \begin{subfigure}[b]{0.2\textwidth}
            \centering
            \includegraphics[width=1.5in]{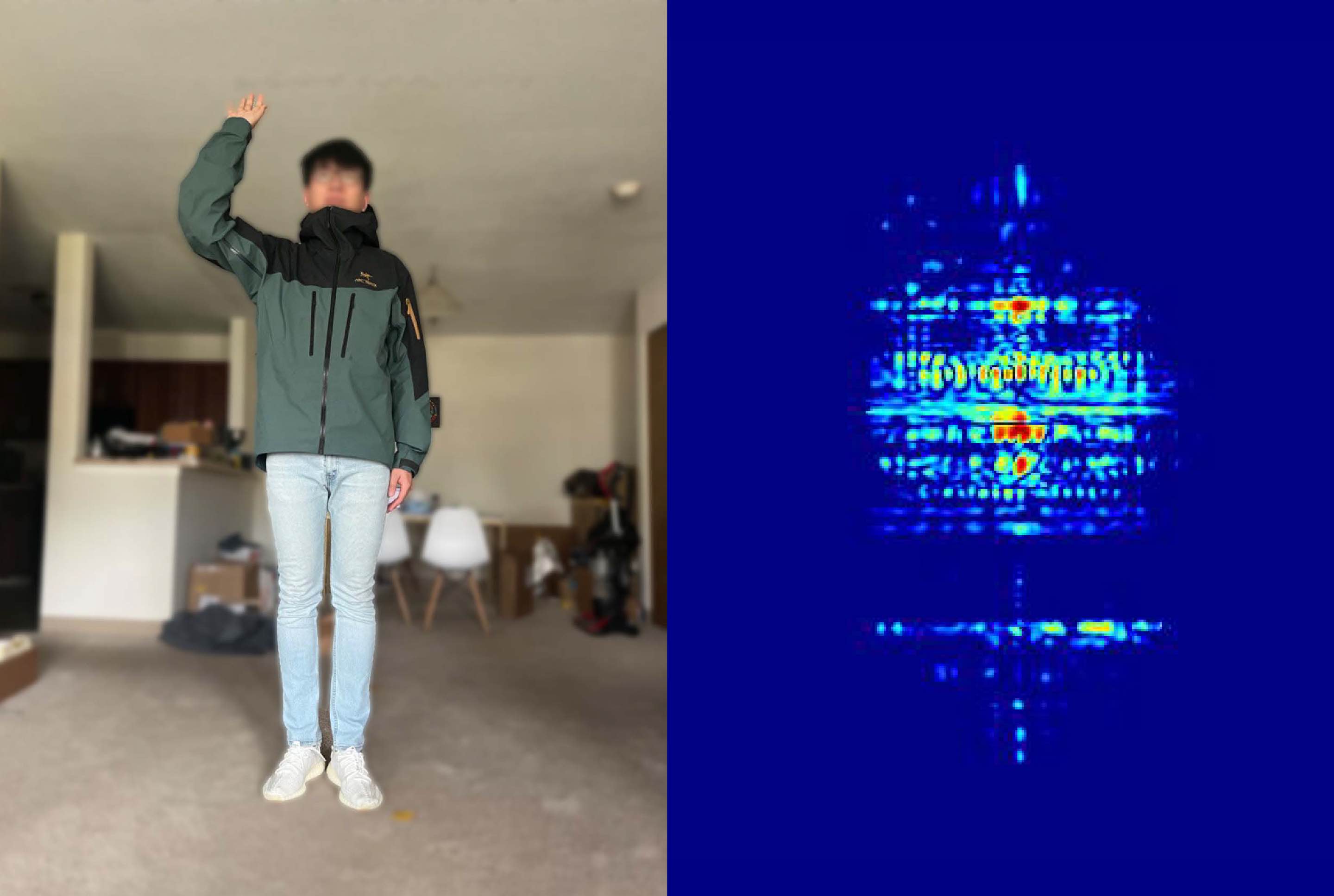}
            \caption{Raising one hand.}
            \label{fig:subfig:raise-one-hand}
        \end{subfigure}
        \hfill
        \begin{subfigure}[b]{0.2\textwidth}
            \centering
            \includegraphics[width=1.5in]{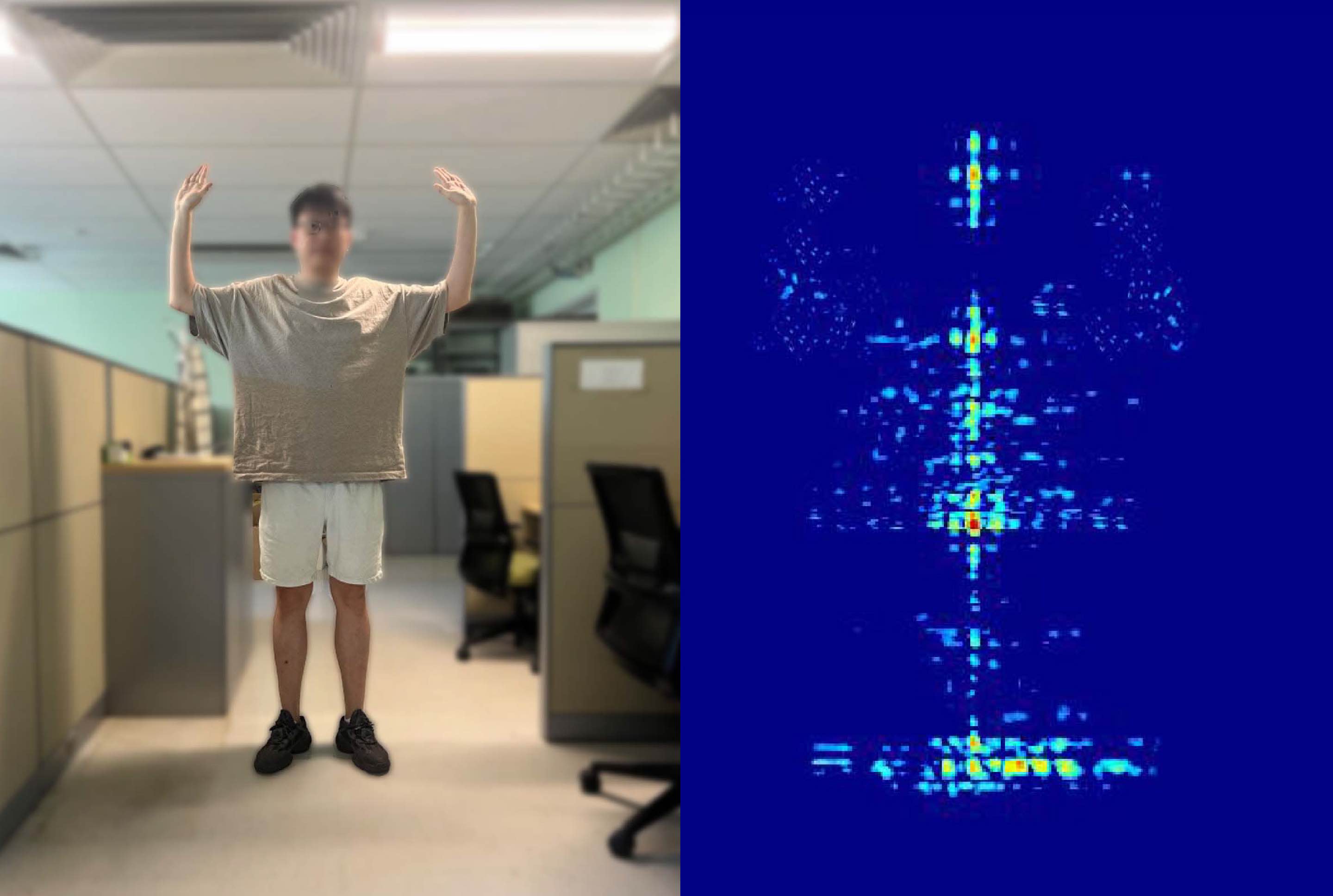}
            \caption{Raising both hands.}
            \label{fig:subfig:raise-two-hand}
        \end{subfigure}
        
        \vspace{0.5cm}
        
        \begin{subfigure}[b]{0.2\textwidth}
            \centering
            \includegraphics[width=1.5in]{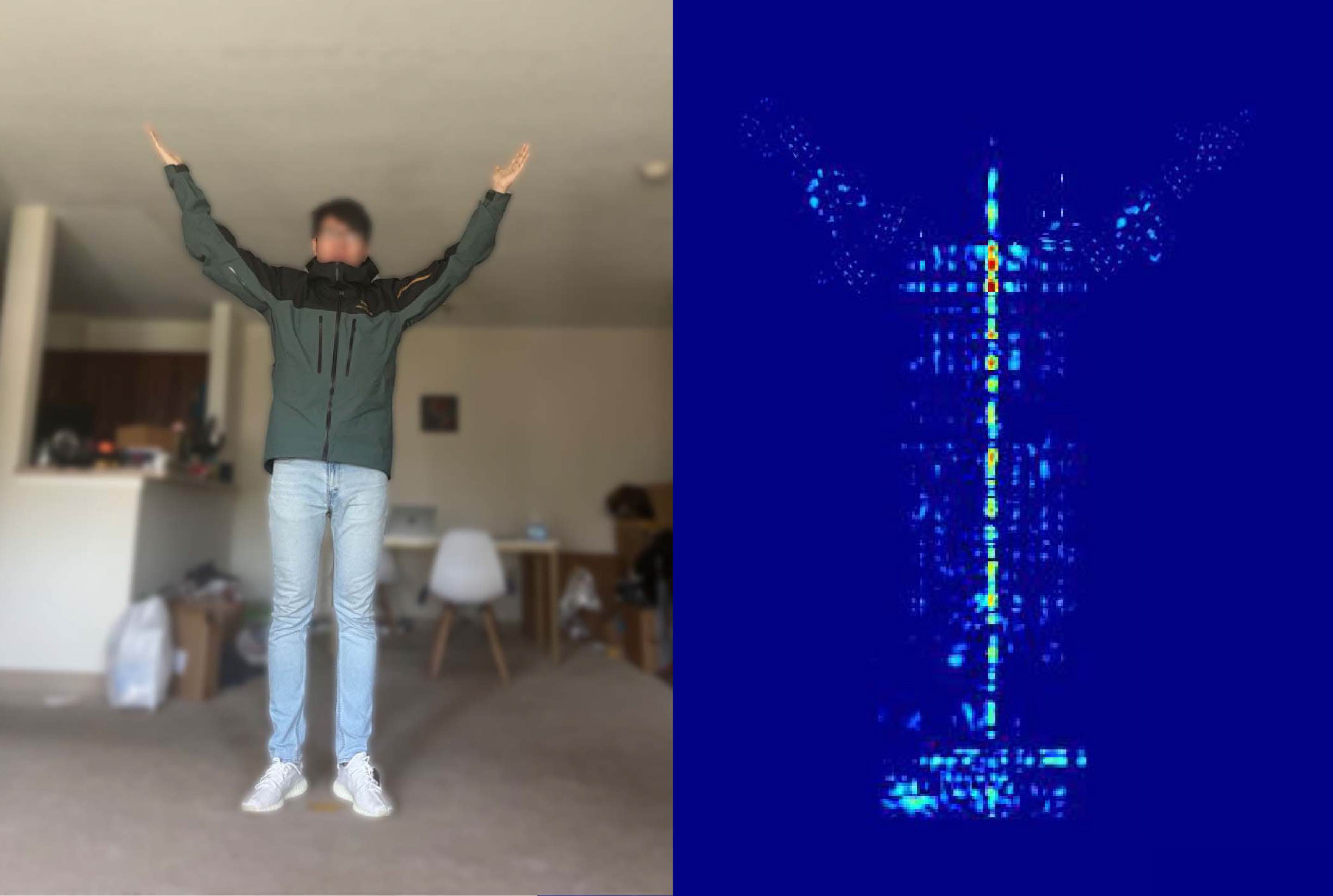}
            \caption{Arm spread diagonally upward.}
            \label{fig:subfig:arm-upward}
        \end{subfigure}
        \hfill
        \begin{subfigure}[b]{0.2\textwidth}
            \centering
            \includegraphics[width=1.5in]{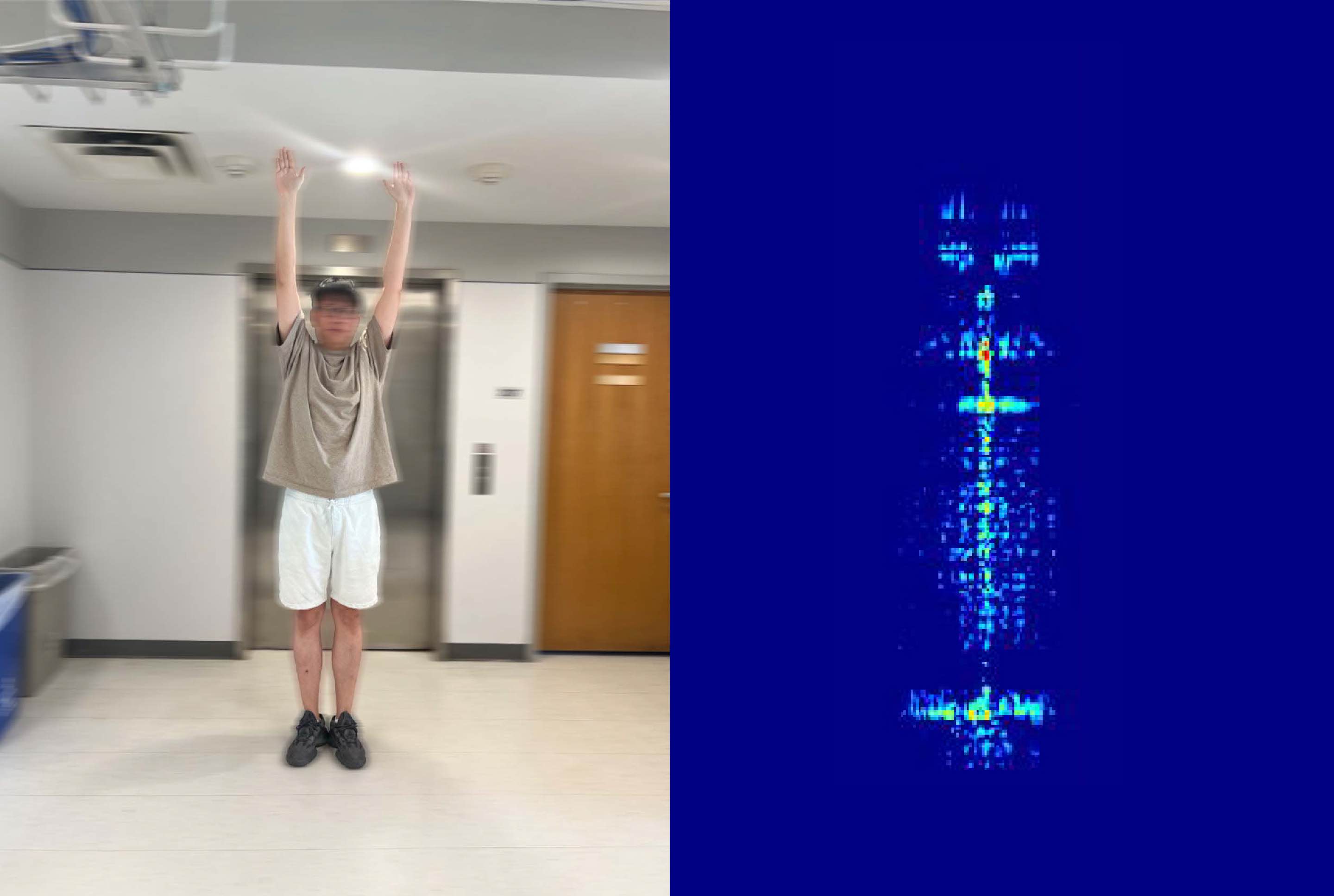}
            \caption{Arms extended straight up.}
            \label{fig:subfig:straight-up}
        \end{subfigure}
        \hfill
        \begin{subfigure}[b]{0.2\textwidth}
            \centering
            \includegraphics[width=1.5in]{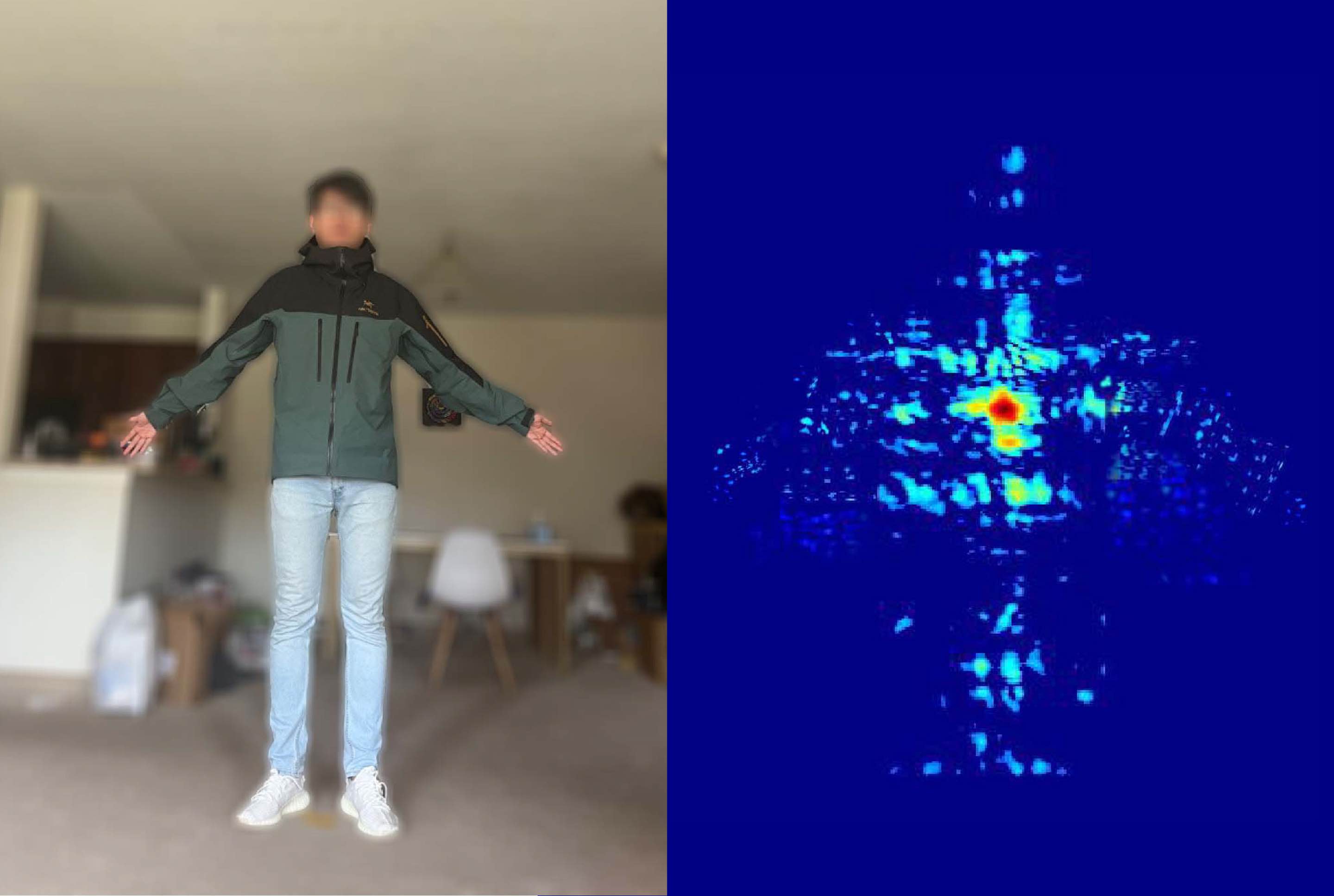}
            \caption{Arms stretched downward.}
            \label{fig:subfig:arm-downward}
        \end{subfigure}
        \hfill
        \begin{subfigure}[b]{0.2\textwidth}
            \centering
            \includegraphics[width=1.5in]{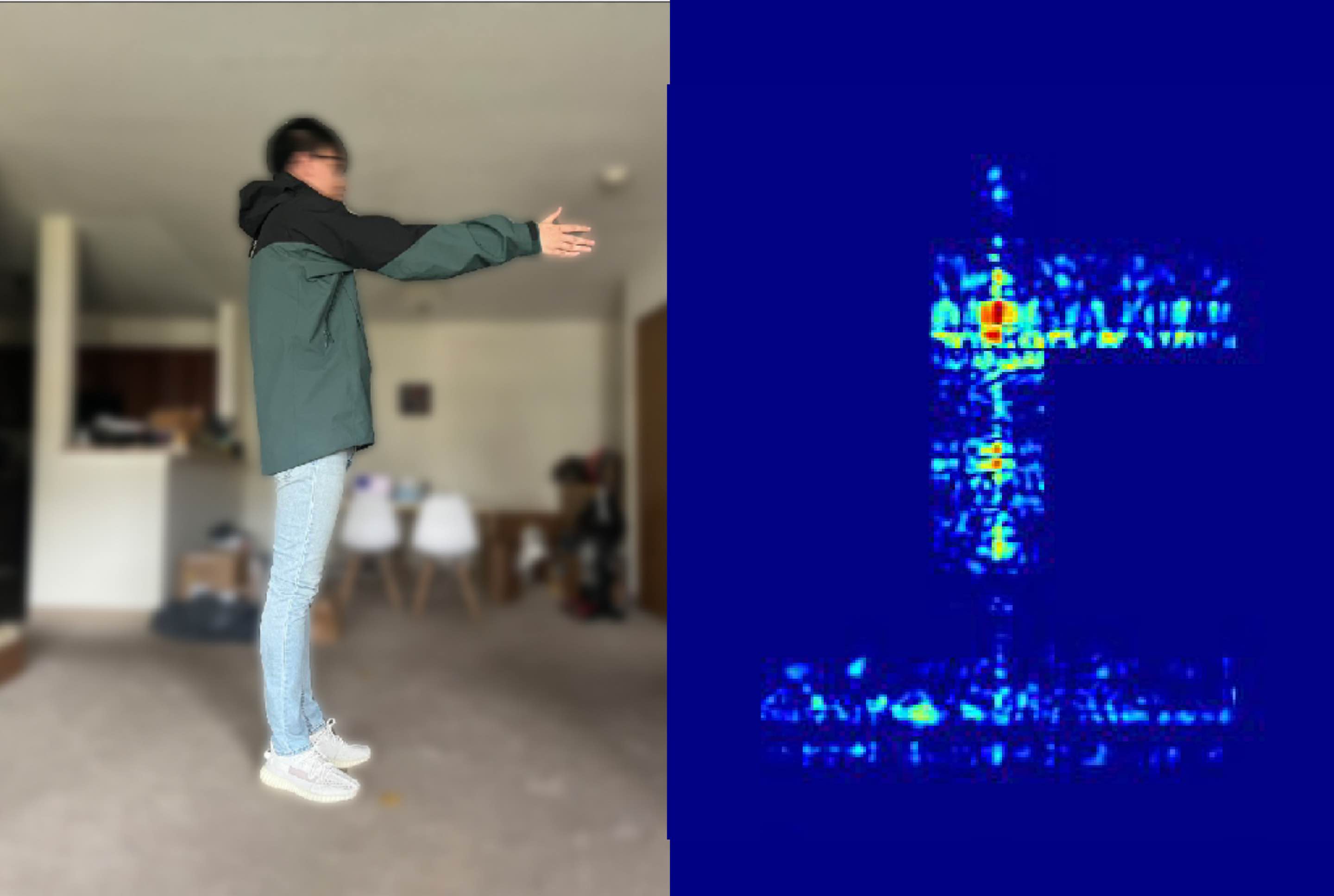}
            \caption{Raising one arm to the side.}
            \label{fig:subfig:raise-one-arm}
        \end{subfigure}
    \end{center}
    \caption{The signal illustration for imaging.}
    \label{fig:actions}
\end{figure}

For our dataset of acoustic images, we defined eight common daily actions: standing, arms spreading, raising one hand, raising both hands, arms spread diagonally upward, arms stretched downward, arms extended straight up, and raising one arm to the side. For each action, data was collected from $20$ volunteers ($10$ male and $10$ female) with $100$ samples per action. We introduced 3 challenges—poor lighting, occlusions, and smoke (to simulate NLOS)—each representing $25\%$ of the total collected data. Fig~\ref{fig:actions} shows these poses and their corresponding generated acoustic images. We used a calibrated Samsung S24 as the camera system to capture RGB images, which serve as the ground truth in our dataset.

%\vfill\pagebreak

% \section{REFERENCES}
% \label{sec:refs}

% List and number all bibliographical references at the end of the
% paper. The references can be numbered in alphabetic order or in
% order of appearance in the document. When referring to them in
% the text, type the corresponding reference number in square
% brackets as shown at the end of this sentence \cite{C2}. An
% additional final page (the fifth page, in most cases) is
% allowed, but must contain only references to the prior
% literature.

\small
\bibliographystyle{IEEEbib}
\bibliography{main}

\end{document}

%% file: preamble.tex
\usepackage{booktabs}
\usepackage{multirow}

\usepackage[T1]{fontenc}
\usepackage[utf8]{inputenc}
\usepackage{graphicx}
\usepackage[font=small]{caption}
\usepackage[font=scriptsize]{subcaption}
\usepackage[english]{babel}
\usepackage{amsmath}
\usepackage{amsthm}

\usepackage{tabularray}
\UseTblrLibrary{diagbox}
\usepackage{pifont}
\usepackage{adjustbox}

%% file: symbol.tex
\usepackage{xcolor}
\definecolor{darkgreen}{rgb}{0,0.5,0}

\usepackage{acro}

\DeclareAcronym{hmr}{
    short = HMR,
    long = Human Mesh Reconstruction,
}
\DeclareMathOperator{\sinc}{sinc}